\documentclass[journal,onecolumn,12pt]{IEEEtran}

\usepackage{booktabs} 
\usepackage{graphicx}
\usepackage{subcaption}
\usepackage{enumitem}
\usepackage{csquotes}
\usepackage{chapterbib}
\usepackage{nameref}
\usepackage[T1]{fontenc}

\usepackage{url}
\usepackage{amsthm}
\theoremstyle{definition}
\newtheorem{definition}{Definition}[]

\begin{document}
\title{Understanding Optical Music Recognition}
%
%
%

\author{Jorge~Calvo-Zaragoza\textsuperscript{*},~\IEEEmembership{University of Alicante} \\
        Jan~Haji\v{c}~Jr.\textsuperscript{*},~\IEEEmembership{Charles University} \\
        Alexander~Pacha\textsuperscript{*},~\IEEEmembership{TU Wien}
\thanks{* Equal contribution}
}

%
%

%

\IEEEpubid{Article published in ACM Computing Surveys, Vol. 53, No. 4, Article 77. DOI: 10.1145/3397499}


\maketitle

\begin{abstract}
For over 50 years, researchers have been trying to teach computers to read music notation, referred to as Optical Music Recognition (OMR). However, this field is still difficult to access for new researchers, especially those without a significant musical background: few introductory materials are available, and furthermore the field has struggled with defining itself and building a shared terminology. In this work, we address these shortcomings by (1) providing a robust definition of OMR and its relationship to related fields, (2) analyzing how OMR inverts the music encoding process to recover the musical notation and the musical semantics from documents, (3) proposing a taxonomy of OMR, with most notably a novel taxonomy of applications. Additionally, we discuss how deep learning affects modern OMR research, as opposed to the traditional pipeline. Based on this work, the reader should be able to attain a basic understanding of OMR: its objectives, its inherent structure, its relationship to other fields, the state of the art, and the research opportunities it affords.
\end{abstract}


%
\IEEEpeerreviewmaketitle

\section{Introduction}
\label{sec:introduction}
Music notation refers to a group of writing systems with which a wide range of music can be visually encoded so that musicians can later perform it. In this way, it is an essential tool for preserving a musical composition, facilitating permanence of the otherwise ephemeral phenomenon of music. In a broad, intuitive sense, it works in the same way that written text may serve as a precursor for speech. Similar to Optical Character Recognition (OCR) technology that has enabled the automatic processing of written texts, reading music notation also invites automation. In an analogy to OCR, the field of Optical Music Recognition (OMR) covers the automation of this task of ``reading'' in the context of music. However, while musicians can read and interpret very complex music scores, there is still no computer system that is capable of doing so.

We argue that besides the technical challenges, one reason for this state of affairs is also that OMR has not defined its goals with sufficient rigor to formulate its motivating applications clearly, in terms of inputs and outputs. Work on OMR is thus fragmented, and it is difficult for a would-be researcher, and even harder for external stakeholders such as librarians, musicologists, composers, and musicians, to understand and follow up on the aggregated state of the art. The individual contributions are formulated with relatively little regard to each other, although less than 500 works on OMR have been published to date. This makes it hard to combine the numerous contributions and use previous work from other researchers, leading to frequent ``reinventions of the wheel.'' The field, therefore, has been relatively opaque for newcomers, despite its clear, intuitive appeal.

One reason for the unsatisfactory state of affairs was a lack of practical OMR solutions: when one is hard-pressed to solve basic subproblems like staff detection or symbol classification, it seems far-fetched to define applications and chain subsystems. However, some of these traditional OMR sub-steps, which do have a clear definition and evaluation methodologies, have recently seen great progress, moving from the category of ``hard'' problems to ``close to solved,'' or at least clearly solvable \cite{Gallego2017, Pacha2017}. Therefore, the breadth of OMR applications that have long populated merely the introductory sections of articles now comes within practical reach. As the field garners more interest within the document recognition and music information retrieval communities \cite{Baro2017, Calvo-Zaragoza2017b, Inesta2018, Paeaekkoenen2018, Achankunju2018, Crawford2018, Rizo2018, Gotham2018, Hajicjr.2018}, we see further need to clarify how OMR talks about itself.

The primary contributions of this paper are to clearly define what OMR is, what problems it seeks to solve and why.\footnote{We assume that readers are familiar with the basics of music notation. While we do revisit some concepts, we recommend Schobrun's step-by-step guide \cite{Schobrun2005} for a thorough introduction into music notation.} OMR is, unfortunately, a somewhat opaque field due to the fusion of the music-centric and document-centric perspectives. Even for researchers, it is difficult to clearly relate their work to the field, as illustrated in Section \ref{sec:WhatIsOpticalMusicRecognition}.

Many authors also think of OMR as notoriously difficult to evaluate \cite{Byrd2015,Hajicjr.2016}. However, we show that this clarity also disentangles OMR tasks which are genuinely hard to evaluate, such as full re-typesetting of the score, from those where established methodologies can be applied straightforwardly, such as searching scenarios.

Furthermore, the separation between music notation as a visual language and music as the information it encodes is sometimes not made clear, which leads to a confusing terminology. The way we formulate OMR should provide a framework of thought in which this distinction becomes obvious.

In order to be a proper tutorial on OMR, this paper addresses certain shortcomings in the current literature, specifically by providing:

\begin{itemize}
\item A robust definition of what OMR is, and a thorough analysis of its inherent structure;
\item Terminological clarifications that should make the field more accessible and easier to survey;
\item A review of OMR uses and applications; well-defined in terms of inputs and outputs, and---as much as possible---recommended evaluation methodologies;
\item A brief discussion of how OMR was traditionally approached and how modern machine learning techniques (namely deep learning) affect current and future research;
\item As supplementary material, an extensive, extensible, accessible, and up-to-date bibliography of OMR (see \nameref{sec:AppendixA}).
\end{itemize}

The novelty of this paper thus lies in collecting and systematizing the fragments found in the existing literature, all in order to make OMR more approachable, easier to collaborate on, and---hopefully---progress faster.

\section{What is Optical Music Recognition?}
\label{sec:WhatIsOpticalMusicRecognition}
So far, the literature on OMR does not really share a common definition of what OMR is. Most authors agree on some intuitive understanding, which can be sketched out as ``computers reading music.'' But until now, no rigorous analysis of this question has been carried out, as most of the literature on the field focuses on providing solutions---or, more accurately, solutions to certain subproblems. These solutions are usually justified by a certain envisioned application or by referencing a review paper that elaborates on common motivations, with Rebelo et al.~\cite{Rebelo2012} being the most prominent one. However, even these review papers \cite{Blostein1992, Bainbridge2001, Rebelo2012, Novotny2015} focus almost exclusively on technical OMR solutions and avoid elaborating the scope of the research.

A critical review of the scientific literature reveals a wide variety of definitions for OMR with two extremes: On one end, the proposed definitions are clearly motivated by the (sub)problem which the authors sought to solve (e.g., ``transforming images of music scores into MIDI files'') which leads to a definition that is too narrow and does not capture the full spectrum of OMR. On the other end, there are some definitions that are so generic that they fail to outline what OMR actually is and what it tries to achieve. An obvious example would be to define OMR as ``OCR for music.'' This definition is overly vague, and the authors are---as likewise in many other papers---particularly unspecific when it comes to clarifying what it actually includes and what not. We have observed that the problem statements and definitions in these papers are commonly adapted to fit the provided solution or to demonstrate the relevance to a particular target audience, e.g., computer vision, music information retrieval, document analysis, digital humanities, or artificial intelligence.

While people rely on their intuition to compensate for this lack of accuracy, we would rather prefer to put an umbrella over OMR and name its essence by proposing the following definition.

\begin{definition} \label{def:omr}
Optical Music Recognition is a field of research that investigates how
to computationally read music notation in documents.
\end{definition}

The first claim of this definition is that OMR is a \emph{research field}. In the published literature, many authors refer to OMR as ``task'' or ``process,'' which is insufficient, as OMR cannot be properly formalized in terms of unique inputs and outputs (as discussed in Section \ref{sec:ATaxonomyOfOmr}). OMR must, therefore, be considered something bigger, like the embracing research field, which investigates how to provide a computer with the ability to read music notation. Within this research field, several tasks can be formulated with specific, unambiguous input-output pairs.

The term ``\emph{computationally}'' distinguishes OMR from the musicological and paleographic studies of how to decode a particular notation system. It also excludes studying how humans read music. OMR does not study the music notation systems themselves---rather, it builds upon this knowledge, with the goal that a computer should be able to read the music notation as well.

The last part of the definition ``\emph{reading music notation in documents}'' tries to define OMR in a concise, clear, specific, and inclusive way. To fully understand this part of the definition, the next section clarifies what kind of information is captured in a music notation document and outlines the process by which it gets generated. The subsequent section then elaborates on how OMR attempts to invert this process to read and recover the encoded information.

It should be noted that the output of OMR is omitted intentionally from its definition, as different tasks require different outputs (see Section \ref{sec:ATaxonomyOfOmr}) and specifying any particular output representation would make the definition unnecessarily restrictive.

\begin{figure}
\includegraphics[width=\textwidth]{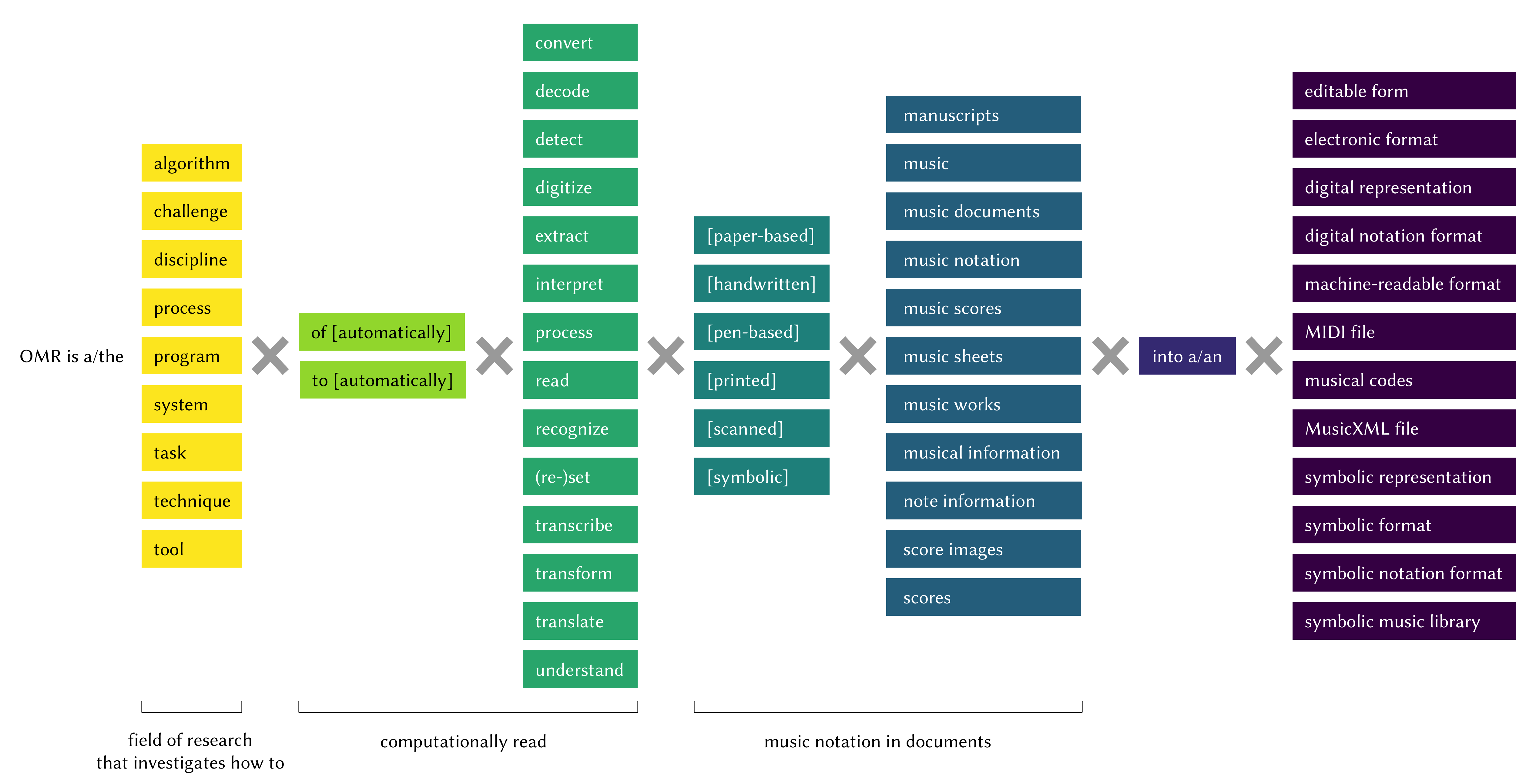}
\caption{How OMR tends to be defined or described and how our proposed definition relates to it. For example: ``OMR is the challenge of (automatically) converting (handwritten) scores into a digital representation.''}
\label{fig:OmrDefinitions}
\end{figure}

To conclude this section, Fig. \ref{fig:OmrDefinitions} illustrates how various definitions of OMR in the literature relate to our proposed definition and are captured by it.

\section{From ``Music'' to a Document}
\label{sec:FromMusicToADocument}
Music can be conceptualized as a structure of \emph{notes in time}. This is not necessarily the only way to conceptualize music,\footnote{As evidenced by either very early music (plainchant) or some later twentieth century compositional styles (mostly spectralism).} but it is the only one that has a consistent, broadly accepted visual language used to transmit it in writing, so it is the conceptualization we consider for the purposes of OMR. A note is a musical object that is defined by four parameters: \emph{pitch, duration, loudness}, and \emph{timbre}. Additionally, it has an \emph{onset}: a placement onto the axis of time, which in music does not mean wall-clock time, but is measured in relative units called beats.\footnote{Musical time is projected onto wall-clock time with an underlying tempo, which can further be stretched and compressed by the performer. Strictly speaking, the notion of beats might not be entirely applicable to some very early music and some contemporary music, where the rhythmic pulse is not clearly defined. However, the notation used to express such music usually does have beats.} Periods of musical time during which no note is supposed to be played are marked by rests, which only have an onset and a duration. Notes and rests can simultaneously be grouped into graphical hierarchies, e.g., multiple eighth notes that are joined by a single beam, as well as logical hierarchies such as phrases or voices. This structure is a vital part of music---it is essential to work it out for making a composition comprehensible.

In order to record this ``conceptualization of music'' visually, for it to be performed over and over in (roughly) the same way, at least at the relatively coarse level of notes, multiple music notation systems have evolved. A music notation system is a visual language that aims to encode music into a graphical form and enrich it with information on \emph{how to perform} it (e.g., bowing marks, fingerings, or articulations).\footnote{Feist \cite{Feist2017} refers to notation whimsically as a ``haphazard Frankenstein soup of tangentially related alphabets and hieroglyphics via which music is occasionally discussed amongst its wonkier creators.''} To do that, it uses a set of symbols as its alphabet and specific rules for how to position these symbols to capture a musical idea. Note that all music notation systems entail a certain loss of information as they are designed to preserve the most relevant properties of the composition very accurately, especially the pitches, durations, and onsets of notes, while under-specifying or even intentionally omitting other aspects.
Tempo could be one of these aspects, where the composer might have expressed precise metronomic indication, given a verbal hint, or stated nothing at all. It is therefore considered the responsibility of the performer to fill those gaps appropriately. We consider this as a natural boundary of OMR: it ends where musicians start to disagree over the same piece of music.

Arguably the most frequently used notation system is \emph{Common Western Music Notation} (CWMN, also known as modern staff notation), which evolved during the seventeenth century from its mensural notation predecessors and stabilized at the beginning of the nineteenth century. There have been many attempts to replace CWMN, but so far, these have not produced workable alternatives. Apart from CWMN, there exist a wealth of modern tablature scores for guitar, used i.e., to write down popular music as well as a significant body of historical musical documents that use earlier notation systems (e.g., mensural notations, quadratic notation for plainchant, early organum, or a wealth of tablature notations for lutes and other instruments).

\begin{figure}
\centering
\begin{subfigure}[b]{\textwidth}
   \includegraphics[width=1\linewidth]{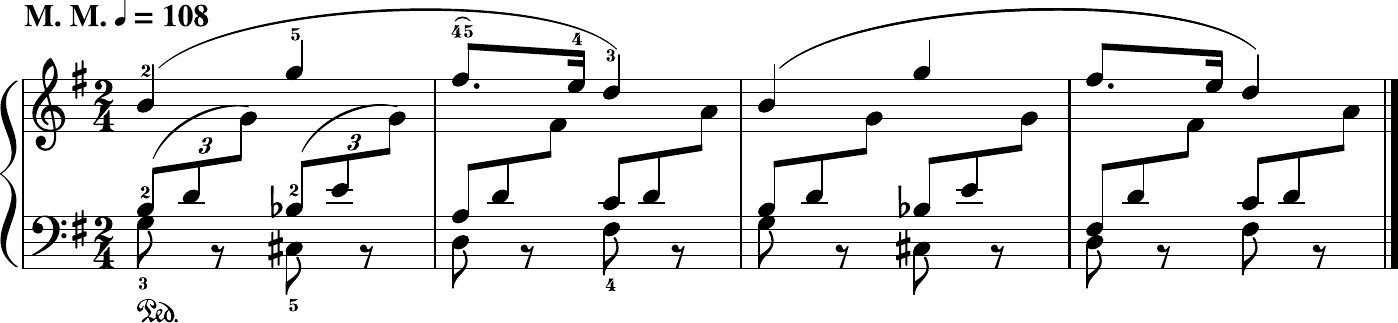}
   \caption{}
\end{subfigure}

\begin{subfigure}[b]{\textwidth}
   \includegraphics[width=1\linewidth]{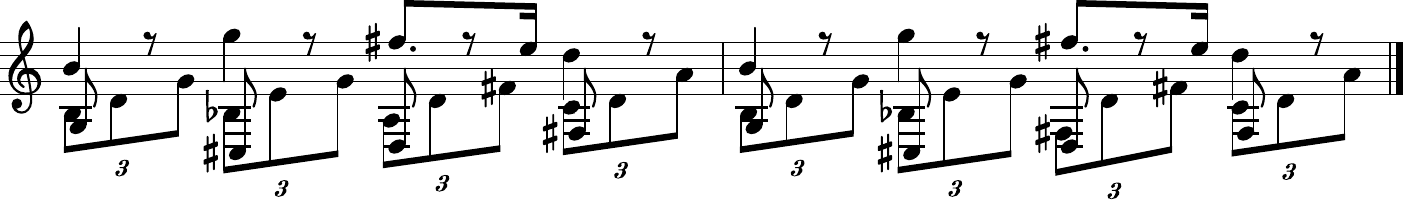}
   \caption{}
\end{subfigure}
\caption{Excerpt of Robert Schumann's ``Kinderszenen'', Op. 15 for piano. Properly engraved (a), it has two staves for the left and the right hand with three visible voices, a key signature and phrase markings to assist the musician. In a poor engraving of the same music (b), that logical structure is lost, and it becomes painfully hard to read and comprehend the music, although these two versions contain the same notes.}
\label{fig:SchumannWithGoodAndBadEngraving}
\end{figure}

Once a music notation system is selected for writing down a piece of music, it is still a challenging task to engrave the music because a single set of notes can be expressed in many ways.\footnote{Normally, music engraving is defined as the process of drawing or typesetting music notation with a high quality for mechanical reproduction. However, we use the term to refer to ``planning the page'': selecting music notation elements and planning their layout to most appropriately capture the music, before it is physically (or digitally) written on the page. This is a loose analogy to the actual engraving process, where the publisher would carefully prepare the printing plates from soft metal, and use them to produce many copies of the music; in our case, this ``printing process'' might not be very accurate, e.g., in manuscripts.} The engraving process involves complex decisions \cite{Blostein1991} that can affect only a local area, like spacings between objects but can also have global effects, like where to insert a page break to make it convenient for the musician to turn the page. For example, one must make sure that the stem directions mark voices consistently and appropriate clefs are used, in order to make the music as readable as possible \cite{Heussenstamm1987, Selfridge-Field1997, Gould2011, Feist2017}. These decisions not only affect the visual appearance but also help to preserve the logical structure (see Fig. \ref{fig:SchumannWithGoodAndBadEngraving}). Afterwards, they can be embodied in a document, whether physically or digitally.

\begin{figure}
\includegraphics[width=\textwidth]{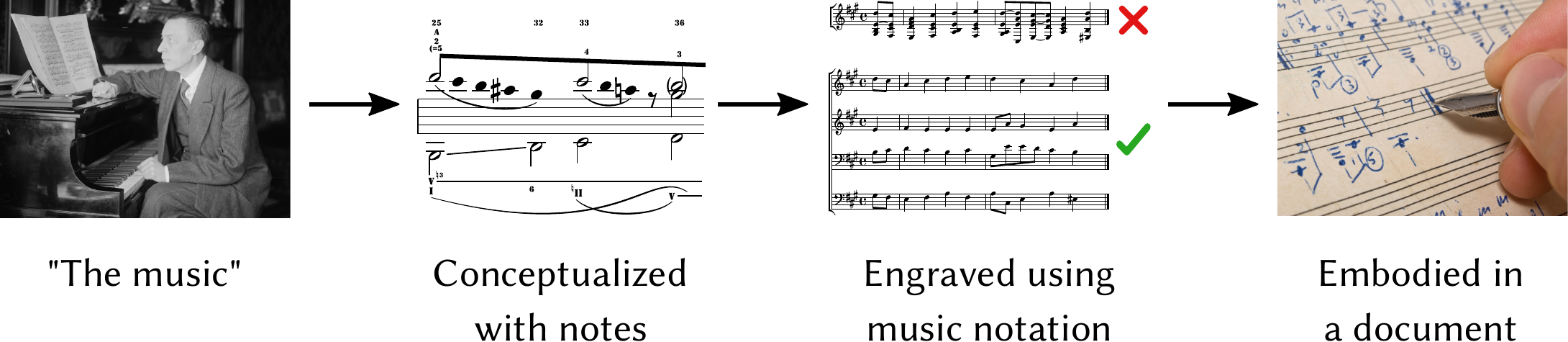}
\caption{How music is typically expressed and embodied (written down).}
\label{fig:CreationOfMusic}
\end{figure}

To summarize, music can be formalized as a structured assembly of notes, enriched through additional instructions for the performer that are encoded visually using a music notational language and embodied in a medium such as paper (see Fig. \ref{fig:CreationOfMusic}). Once this embodiment is digitized, OMR can be understood in terms of inverting this process.

\section{Inverting the Music Encoding Process}
\label{sec:InvertingTheMusicEncodingProcess}
OMR starts after a musical composition has been expressed visually with music notation in a document.\footnote{In this context, a document is usually understood as a digital image. See Section \ref{sec:OmrInputs} for alternatives.} The music notation document serves as a medium, designed to encode and transmit a musical idea from the composer to the performer, enabling the recovery and interpretation of that envisioned music by reading through it. The performer would:

\begin{enumerate}
\item \emph{Read the visual signal} to determine what symbols are present and what is their configuration,
\item Use this information to \emph{parse and decode the notes and their accompanying instructions} (e.g., indications of which technique to use), and
\item Apply musical intuition, prior knowledge, and taste to \emph{interpret the music} and fill in the remaining parameters which music notation did not capture.
\end{enumerate}

\begin{figure}
\includegraphics[width=\textwidth]{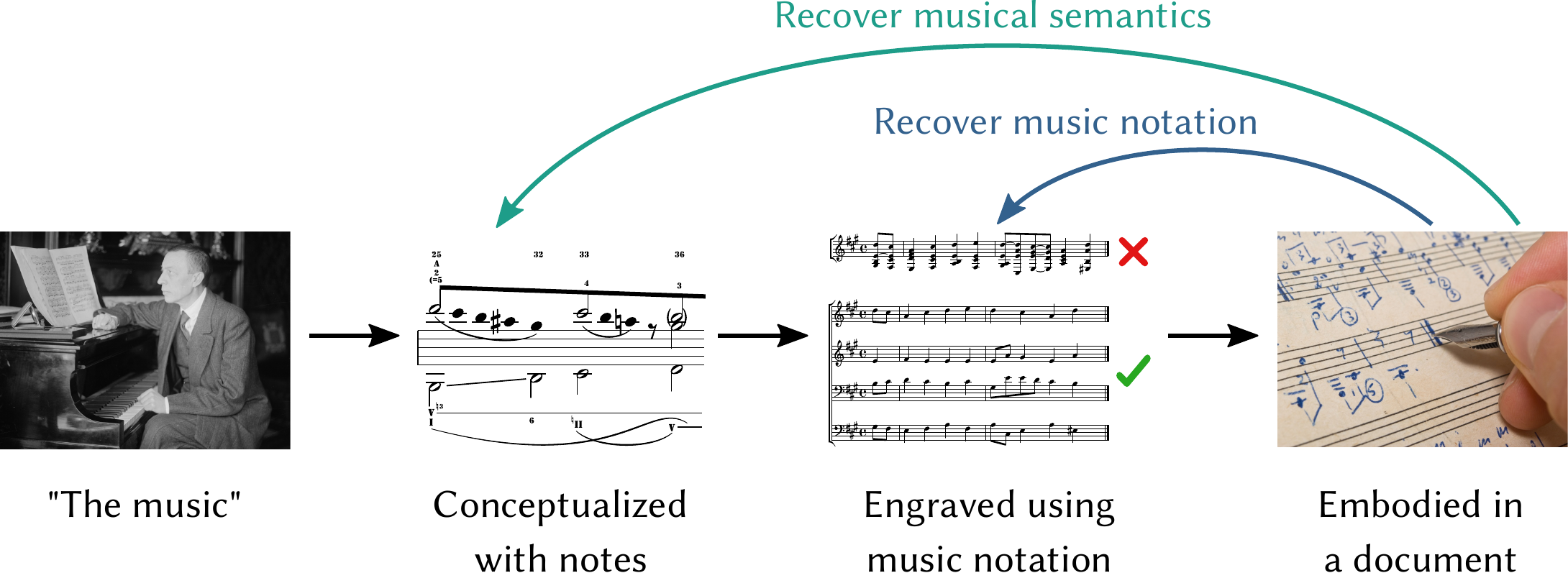}
\caption{How ``reading'' music can be interpreted as the operations of inverting the encoding process.}
\label{fig:InvertingTheCreationProcess}
\end{figure}

Note that step (3) is clearly outside of OMR since it needs to deal with information that is not written into the music document---and where human performers start to disagree, although they are reading the very same piece of music \cite{Lopresti2002}.\footnote{Analogously, speech synthesis is not considered a part of optical character recognition. However, there exists expressive performance rendering software that attempts to simulate more authentic playback, addressing step (3) in our analysis. More information can be found in \cite{Cancino-Chacon2018}.}
Coming back to our definition of OMR, based on the stages of the writing/reading process we outlined above, there are two fundamental ways to interpret the term ``read'' in \emph{reading music notation} as illustrated in Fig. \ref{fig:InvertingTheCreationProcess}. We may wish to:

\begin{enumerate}[label=(\Alph*)]
\item \emph{Recover music notation} and information from the engraving process, i.e. what elements were selected to express the given piece of music and how they were laid out. This corresponds to stage (1) in the analysis above and does not necessarily require specific musical knowledge, but it does require an output representation that is capable of storing music notation, e.g., MusicXML or MEI, which can be quite complex.
\item \emph{Recover musical semantics}, which we define as the notes, represented by their pitches, velocities, onsets, and durations. This corresponds to stage (2)---we use the term ``semantics'' to refer only to the information that can be unambiguously inferred from the music notation document. In practical terms, MIDI would be an appropriate output representation for this goal.
\end{enumerate}

This is a fundamental distinction that dictates further system choices, as we discuss in the next sections. Note that, counter-intuitively, going backwards through this process just one step (A - recover music notation) might be in fact more difficult than going back two steps (B - recover musical semantics) directly. This is because music notation contains a logical structure and more information than simply the notes. Skipping the explicit description of music notation allows this complexity to be bypassed.

There is, of course, a close relationship between recovering music notation and musical semantics. A single computer program may even attempt to solve both at the same time because once the full score with all its notational details is recovered, the musical semantics can be inferred unambiguously. Keep in mind that the other direction does not necessarily work: if only the musical semantics are restored from a document without the engraving information that describes how the notes were arranged, those notes may still be typeset using meaningful engraving defaults, but the result is probably much harder to comprehend (see Fig. \ref{fig:SchumannWithGoodAndBadEngraving}b for such an example).

\subsection{Alternative Names}
\label{sec:AlternativeNames}
\emph{Optical Music Recognition} is a well-established term, and we do not seek to establish a new one. We just notice a lack of precision in its definition. Therefore, it is no wonder that people have been interpreting it in many different ways to the extent that even the optical detection of lip motion for identifying the musical genre of a singer \cite{Ding2014} has been called OMR. Alternative names that might not exhibit this vagueness are Optical Music Notation Recognition, Optical Score Recognition, or Optical Music Score Recognition.

\section{Relation To Other Fields}
\label{sec:RelationToOtherFields}
Now that we have thoroughly described what \emph{Optical Music Recognition} is, we briefly set it in context of other disciplines, both scientific and general fields of human endeavors.

\begin{figure}[ht]
\centering
\includegraphics[width=0.65\textwidth]{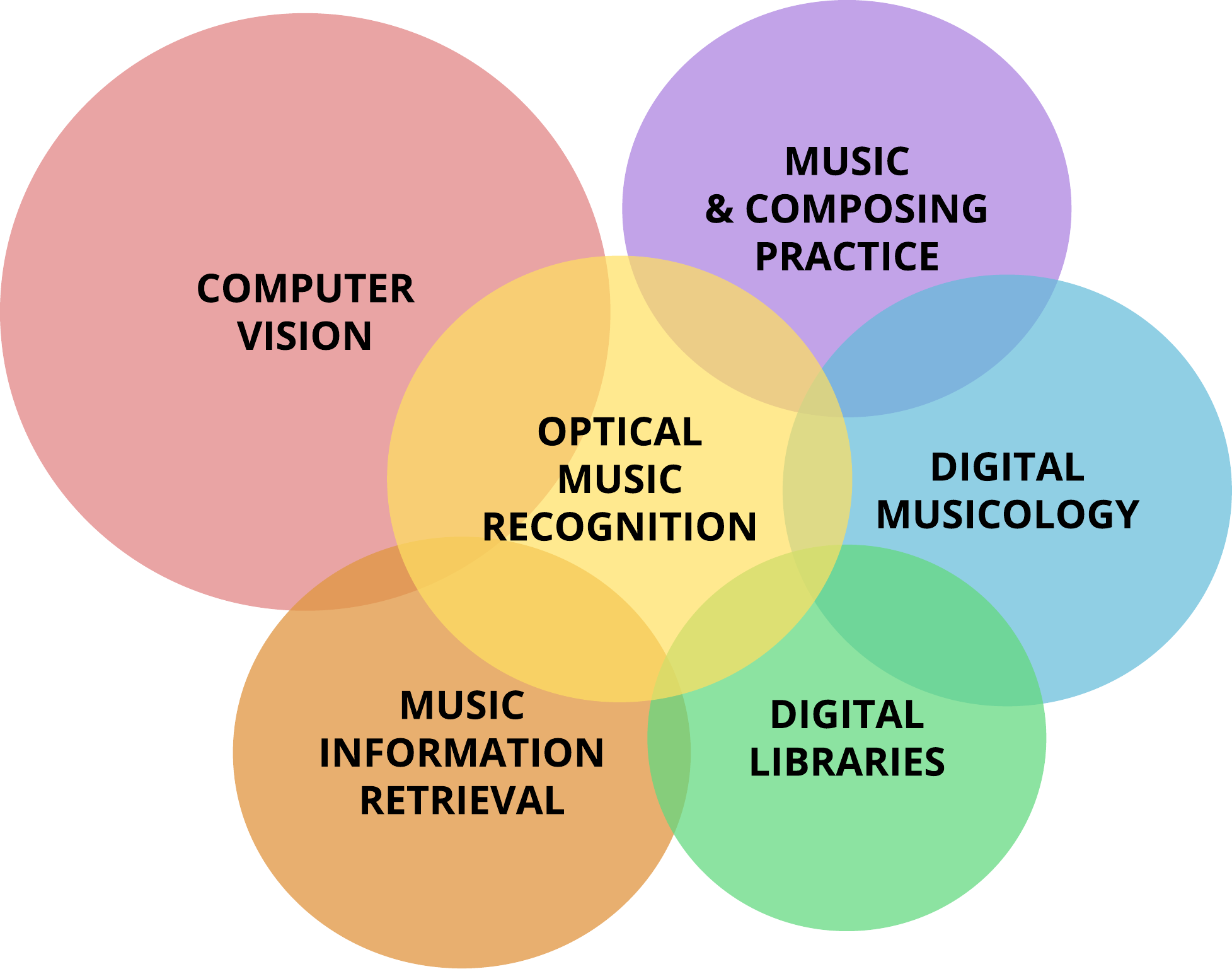}
\caption{Optical Music Recognition with its most important related fields, methods, and applications.}
\label{fig:OmrRelations}
\end{figure}

Figure \ref{fig:OmrRelations} lays out the various key areas that are relevant for OMR, both as its tools and the ``consumers'' of its outputs. From a technical point of view, OMR can be considered a subfield of computer vision and document analysis, with deep learning acting as a catalyst that opens up promising novel approaches. Within the context of Music Information Retrieval (MIR), OMR should enable the application of MIR algorithms that rely on symbolic data and audio inputs (through rendering the recognized scores). It furthermore can enrich digital music score libraries and make them much more searchable and accessible, which broadens the scope of digital musicology to compositions for which we only have the written score (which is probably the majority of Western musical heritage). Finally, OMR has practical implications for composers, conductors, and the performers themselves, as it cuts down the costs of digitizing scores, and therefore bring the benefits of digital formats to their everyday practice.

\subsection{Optical Music Recognition vs. Text Recognition}
One must also address the obvious question: why should OMR be singled out besides Optical Character Recognition (OCR) and Handwritten Text Recognition (HTR), given that they are tightly linked \cite{Bellini2001}, and OMR has frequently been called ``OCR for music'' \cite{Pugin2007b, Burgoyne2008, Jones2008, Johansen2009, Gozzi2010, Pugin2013, Ng2014, Fujinaga2014a, Burgoyne2015, Sotoodeh2017}?\footnote{Even the English Wikipedia article on OMR has been calling it ``Music OCR'' for over 13 years.} What is the justification of talking specifically about music notation and what differentiates it from other graphics recognition challenges? What are the special considerations in OMR that one does not encounter in other writing systems?

A part of the justification lies in the properties of music notation as a \emph{contextual} writing system. While its alphabet consists of well-defined primitives (e.g., stems, noteheads, or flags) that have a clear interpretation, it is only in their configuration---how they are placed and arranged on the staves, and with respect to each other---that specifies what notes should be played. The properties of music notation that make it a challenge for computational reading have been discussed exhaustively by Byrd and Simonsen \cite{Byrd2015}; we hypothesize that these difficulties are ultimately caused by this contextual nature of music notation.

\begin{figure}
\includegraphics[width=\textwidth]{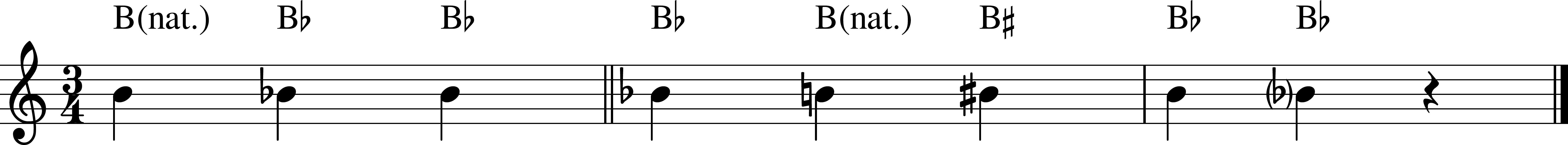}
\caption{How the translation of the graphical concept of a note into a pitch is affected by the clef and accidentals. The effective pitch is written above each note. Accidentals immediately before a note propagate to other notes within the same measure, but not to the next measure. Accidentals at the beginning of a measure indicate a new key signature that affects all subsequent notes.}
\label{fig:Pitch}
\end{figure}

Another major reason for considering the field of OMR distinct from text recognition is the application domain itself---music. When processing a document of music notation, there is a natural requirement to recover its musical semantics as well (see Section \ref{sec:InvertingTheMusicEncodingProcess}, setting B), as opposed to text recognition, which typically does not have to go beyond recognizing letters or words and ordering them correctly. There is no proper equivalent of this interpretation step in text recognition since there is no definite answer to \emph{how a symbol configuration (=words) should be further interpreted}; therefore, one generally leaves interpretation to humans or to other well-defined tasks from the Natural Language Processing field. However, given that music is overwhelmingly often conceptualized as notes, and notes are well-defined objects that can be inferred from the score, OMR is, not unreasonably, asked to produce this additional level of outputs that text recognition does not. Perhaps the simplest example to illustrate this difference is given by the concept of the pitch of the notes (see Fig. \ref{fig:Pitch}). While graphically a note lies on a specific vertical position of the staff, other objects, such as the clefs and accidentals, determine its musical pitch. It is therefore insufficient for the OMR to provide just the results in terms of positions, but it also has to take the context into account, in order to convert positions (graphical concept) into pitches (musical concept). To make matters worse, the correct interpretation depends on conventions, which can change depending on social, historical, or professional context (what a jazz musician accepts as conventional will differ from that understood by a classical player, for example). In this regard, OMR is more ambitious than text recognition, since there is an additional interpretation step specifically for music that has no good analogy in other natural languages.

\begin{figure}
\includegraphics[width=1.0\textwidth]{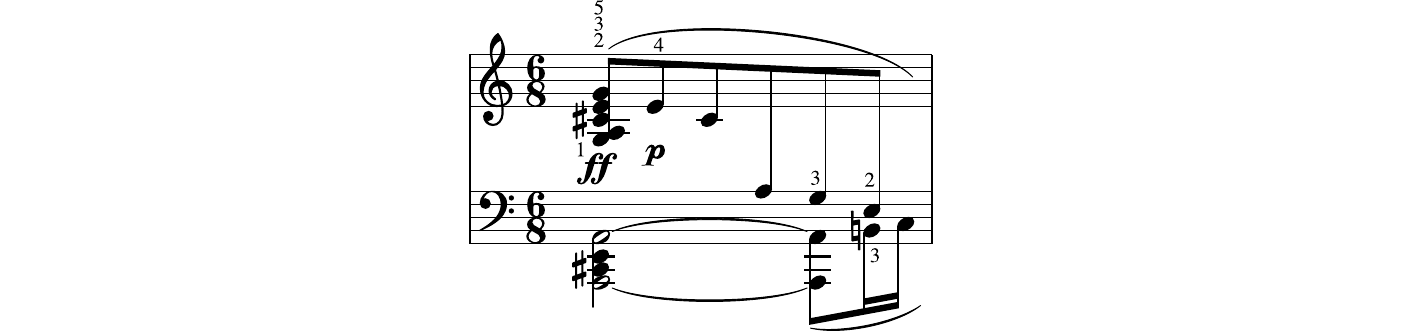}
\caption{This excerpt by Ludwig van Beethoven, Piano Sonata op. 2 no. 2, Largo appassionato, m. 31 illustrates some properties of the music notation that distinguish it from other types of writing systems: a wide range of primitive sizes, the same primitives appearing at different scales and rotations, and the ubiquitous two-dimensional spatial relationships.}
\label{fig:MusicalFeatures}
\end{figure}

The character set poses another significant challenge, compared to text recognition. Although writing systems like Chinese have extraordinarily complex character sets, the primitives in music notation have a much greater variability in size, ranging from small elements like a \emph{dot} to big elements spanning an entire page like the \emph{brace}. Many of the primitives may appear at various scales and rotations like \emph{beams} or have a nearly unrestricted appearance like \emph{slurs} that are only defined as more-or-less smooth curves that may be interrupted anywhere. Finally, in contrast to text recognition, music notation involves ubiquitous two-dimensional spatial relationships, which are salient for the symbols' interpretation. Some of these properties are illustrated in Fig. \ref{fig:MusicalFeatures}.

Furthermore, Byrd and Simonsen \cite{Byrd2015} argue that because of the vague limits of what one may want to express using music notation, its syntactic rules can be expected to be bent accordingly; this happens to such an extent that Homenda et al. \cite{Homenda1996} argued that there is no universal definition of music notation at all. Figure \ref{fig:MusicalFeatures} actually contains an instance of such rule-breaking: while one would expect all notes in one chord to share the same duration, the chord on the bottom left contains a mix of white and black noteheads, corresponding to half- and quarter-notes. At the same time, however, the musical intent is yet another: the two quarter-notes in the middle of the chord are actually played as eighth notes, to add to the rich sonority of the fortissimo chord on the first beat.\footnote{This effect would be especially prominent on the Hammerklavier instruments prevalent around the time Beethoven was composing this sonata.} We believe this example succinctly illustrates the intricacies of the relationship between musical comprehension and music notation. This last difference between a written quarter and interpreted eighth note is, however, beyond what one may expect OMR to do, but it serves as further evidence that the domain of music presents its own difficulties, compared to the domains where text recognition normally operates.

\subsection{Optical Music Recognition vs. Other Graphics Recognition Challenges}
\label{sec:OpticalMusicRecognitionVsOtherGraphicsRecognitionChallenges}
Apart from text, documents can contain a wide range of other graphical information, such as engineering drawings, floor plans, mathematical expressions, comics, maps, patents, diagrams, charts or tables \cite{Chhabra1998, Fornes2018}. Recognizing any of these comes with its own set of challenges, e.g., comics combine text and other visual information in order to narrate a story, which makes recovering the correct reading order a non-trivial endeavor. Similarly, the arrangement of symbols in engineering drawing and floor plans can be very complex with rather arbitrary shapes. Even tasks that are seemingly easy, such as the recognition of tables, must not be underestimated and are still subject to ongoing research \cite{Shahab2010, Rashid2017}. The hardest aspects of OMR are much closer to these challenges than to text recognition: the ubiquitous two-dimensionality, long-distance spatial relationships, and the permissive way of how individual elements can be arranged and appear at different scales and rotations.

\begin{figure}
\centering
\includegraphics[width=0.8\textwidth]{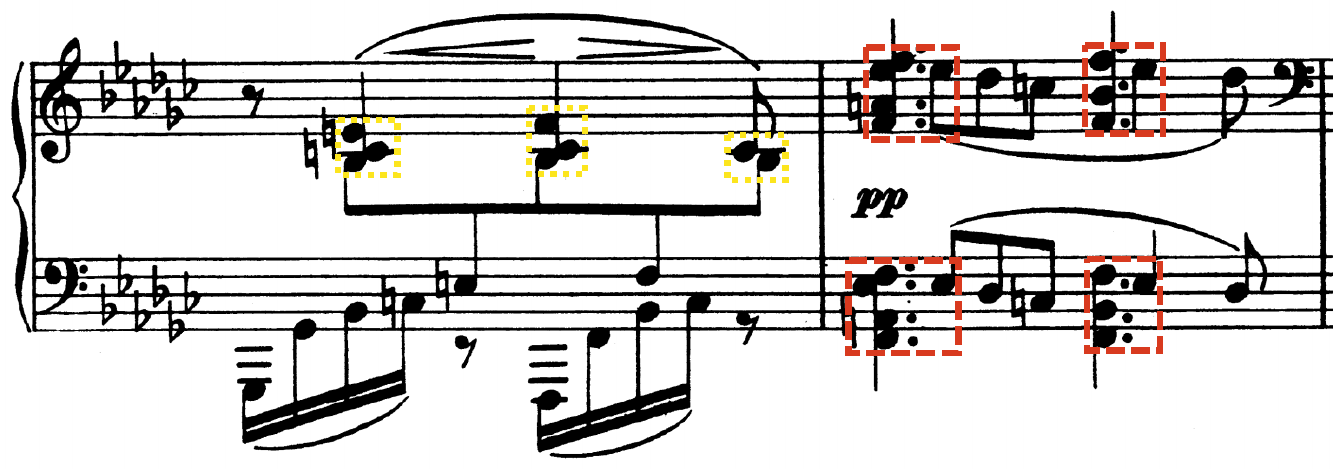}
\caption{Brahms Intermezzo, Op. 117 no. 1. Adjacent notes of the chords in the first bar in the top staff are shifted to the right to avoid overlappings (yellow dotted boxes). The moving eighths in the second bar are forced even further to the right, although being played simultaneously with the chord (red dashed boxes).}
\label{fig:Brahms}
\end{figure}

\begin{figure}
\centering
\includegraphics[width=0.8\textwidth]{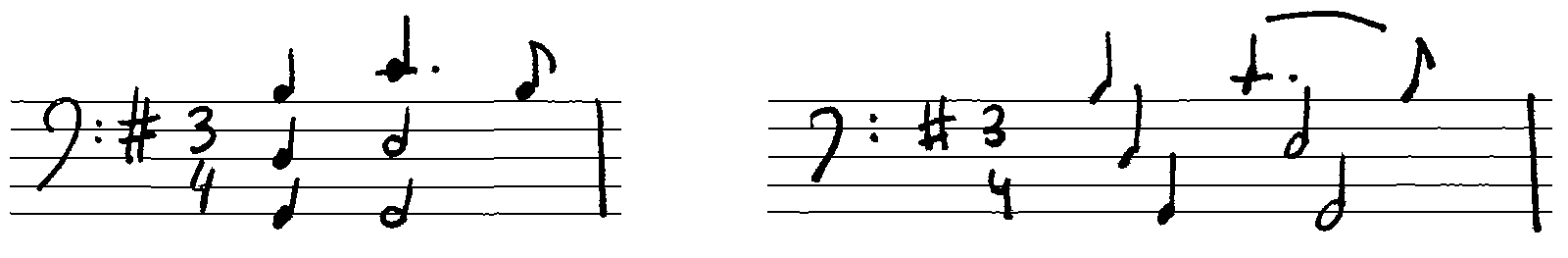}
\caption{Sample from the CVC-MUSCIMA dataset \cite{Fornes2012} with the same bar transcribed by two different writers. The first three notes and the second three notes form a chord and should be played simultaneously (left figure) but sometimes scores are written in a poor musical orthography (right figure) which can create the impression that the notes should be played after each other instead of together.}
\label{fig:MUSCIMA}
\end{figure}

\begin{figure}
\centering
\includegraphics[width=0.6\textwidth]{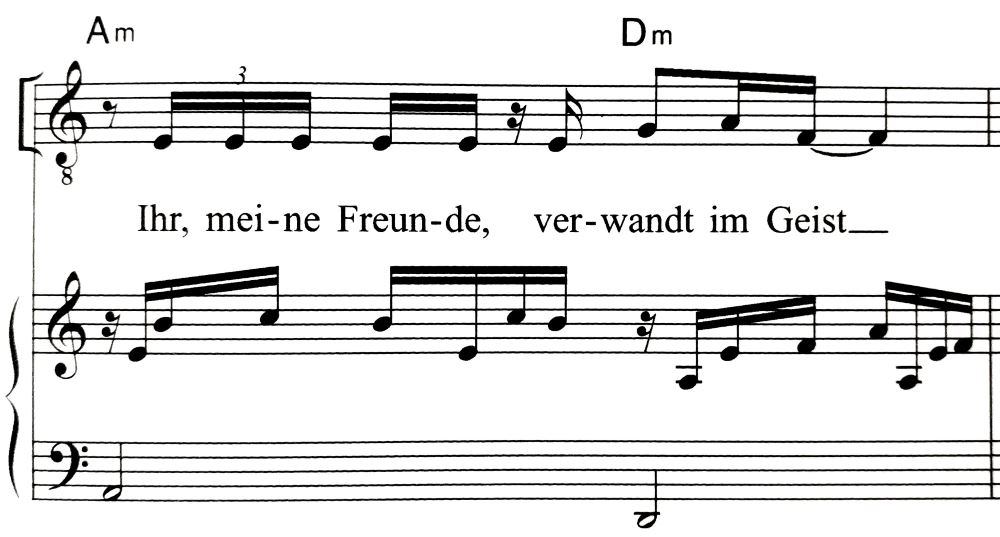}
\caption{Sample from the Songbook of Romeo \& Julia by Gerard Presgurvic \cite{Presgurvic2005} with uneven spacing between multiple sixteenth notes in the middle voice to align them with the lyrics.}
\label{fig:Romeo}
\end{figure}

One thing that makes CWMN more complex than many graphics recognition challenges like mathematical formulae recognition is the complex typographical alignment of objects \cite{Bainbridge2001, Byrd2015} that is dictated by the content, e.g., simultaneous events should have the same horizontal position, or each space between multiple notes of the same length should be equal. Interactions between individual objects can force other elements to move around, potentially breaking those principles (see Fig. \ref{fig:Brahms}, \ref{fig:MUSCIMA} and \ref{fig:Romeo}).

Apart from the typographical challenges, music notation also has extremely complex semantics, with many implicit rules. To handle this complexity, researchers started a long time ago to leverage the rules that govern music notation and formulate them into grammars \cite{Prerau1971, Andronico1982}. For instance, the fact that the note durations (in each notated voice) have to sum up to the length of a measure has been integrated into OMR as a post-processing step \cite{Padilla2014}. Fujinaga \cite{Fujinaga1988} even states that music notation can be recognized by an LL(k) grammar. Nevertheless, the following citation from Blostein and Baird \cite{Blostein1992} (p.425) is still mostly true:

\begin{displayquote}
``Various methods have been suggested for extending grammatical methods which were developed for one-dimensional languages. While many authors suggest using grammars for music notation, their ideas are only illustrated by small grammars that capture a tiny subset of music notation.'' \cite{Blostein1992} (p.425; sec. 7 - Syntactic Methods).
\end{displayquote}

There has been progress on enlarging the subset of music notation captured by these grammars, most notably in the DMOS system \cite{Coueasnon1994}, but there are still no tractable 2-D parsing algorithms that are powerful enough for recognizing music notation without relying on fragile segmentation heuristics. It is not clear whether current parsers used to recognize mathematical expressions \cite{Alvaro2016} are applicable to music notation or simply have not been applied yet---at least we are not aware of any such works.

\section{A Taxonomy of OMR}
\label{sec:ATaxonomyOfOmr}
Now that we have progressed in our effort to define \emph{Optical Music Recognition}, we can turn our attention to systematizing the field with respect to motivating applications, subtasks, and their interfaces. We reiterate that our objective is not to review the methods by which others have attempted to reach the goals of their OMR work; rather, we are proposing a taxonomy of the field's goals themselves. Our motivation is to find natural groups of OMR applications and tasks for which we can expect, among other things, shared evaluation protocols. The need for such systematization has long been felt \cite{Blostein1992a, Calvo-Zaragoza2018d}, but subsequent reviews \cite{Rebelo2012, Novotny2015} have focused almost entirely on technical solutions.

\subsection{OMR Inputs}
\label{sec:OmrInputs}
The taxonomy of \emph{inputs} of OMR systems is generally established. The first fundamental difference can be drawn between \emph{offline} and \emph{online} OMR:\footnote{Although it might sound ambiguous, the term ``online recognition'' has been used systematically in the handwritten recognition community \cite{Plamondon2000}. Sometimes, this scenario is also referred to as pen-based recognition.} offline OMR operates on a static image, while online OMR processes music notation as it is being written, e.g., by capturing the stylus input on an electronic tablet device \cite{George2003, George2004c, Taubman2005, Calvo-Zaragoza2014}. Online OMR is generally considered easier since the decomposition into strokes provides a high-quality over-segmentation essentially for free. Offline OMR can be further subdivided by the engraving mechanism that has been used, which can be either \emph{typeset} by a machine, often inaccurately referred to as \emph{printed},\footnote{Handwritten manuscripts can also be printed out, if they were scanned previously, therefore we prefer the word typeset.} or \emph{handwritten} by a human, with an intermediate, yet common scenario of handwritten notation on pre-printed staff paper. Throughout the rest of the paper, we mainly elaborate on the case of offline OMR, because it represents the most common and considered scenario.

\begin{figure}
    \centering
    \begin{tabular}{cc}
      \includegraphics[width=65mm]{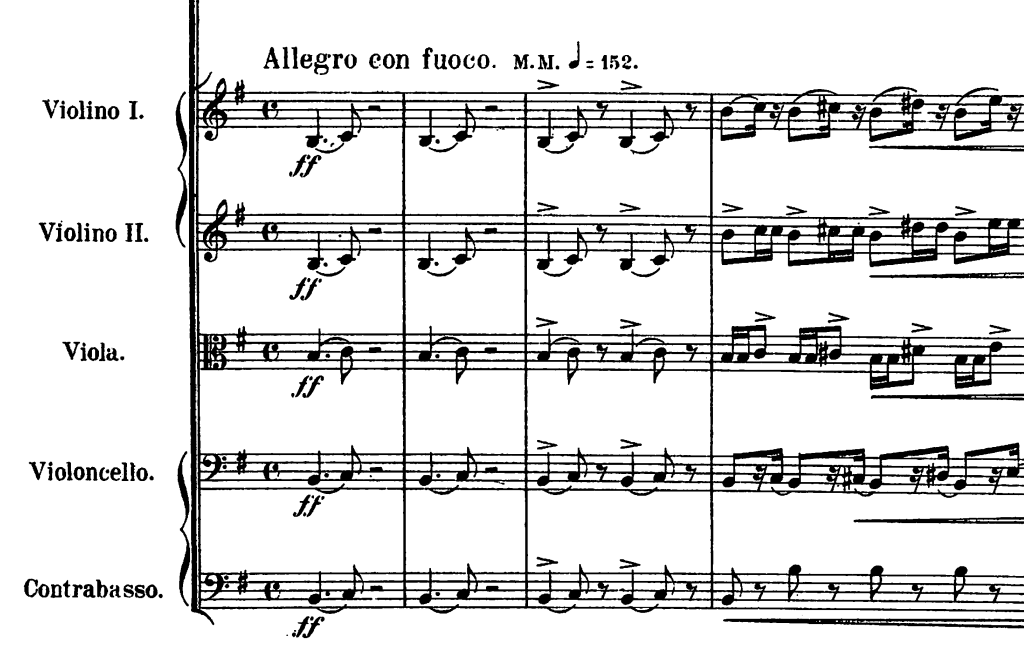} & \includegraphics[width=65mm]{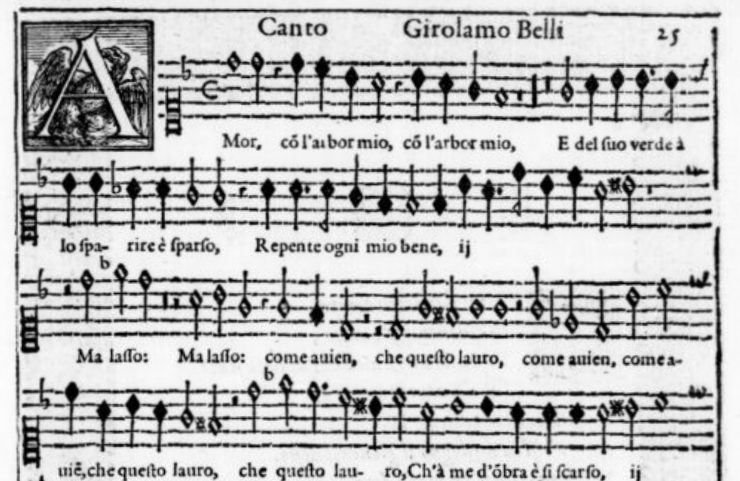} \\
      (a) & (b) \\[10pt]
     \includegraphics[width=65mm]{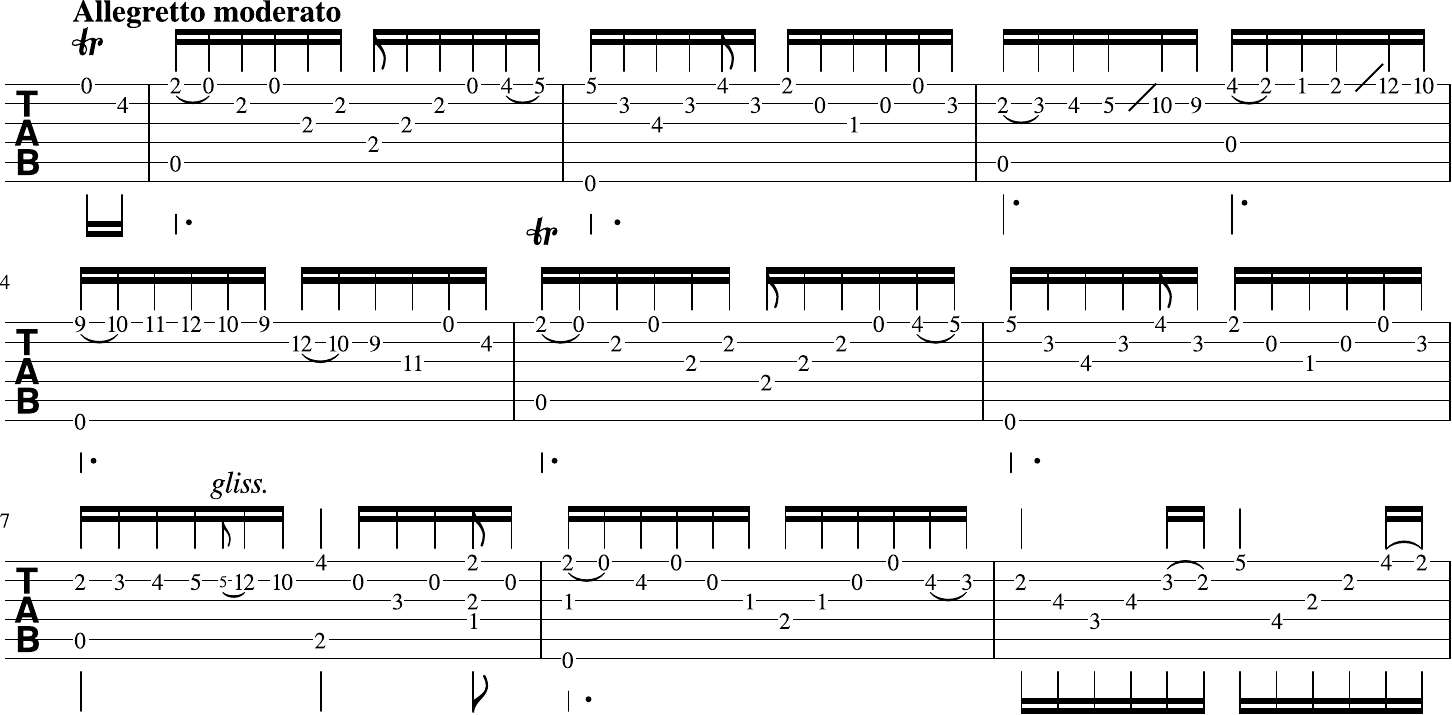} & \includegraphics[width=65mm]{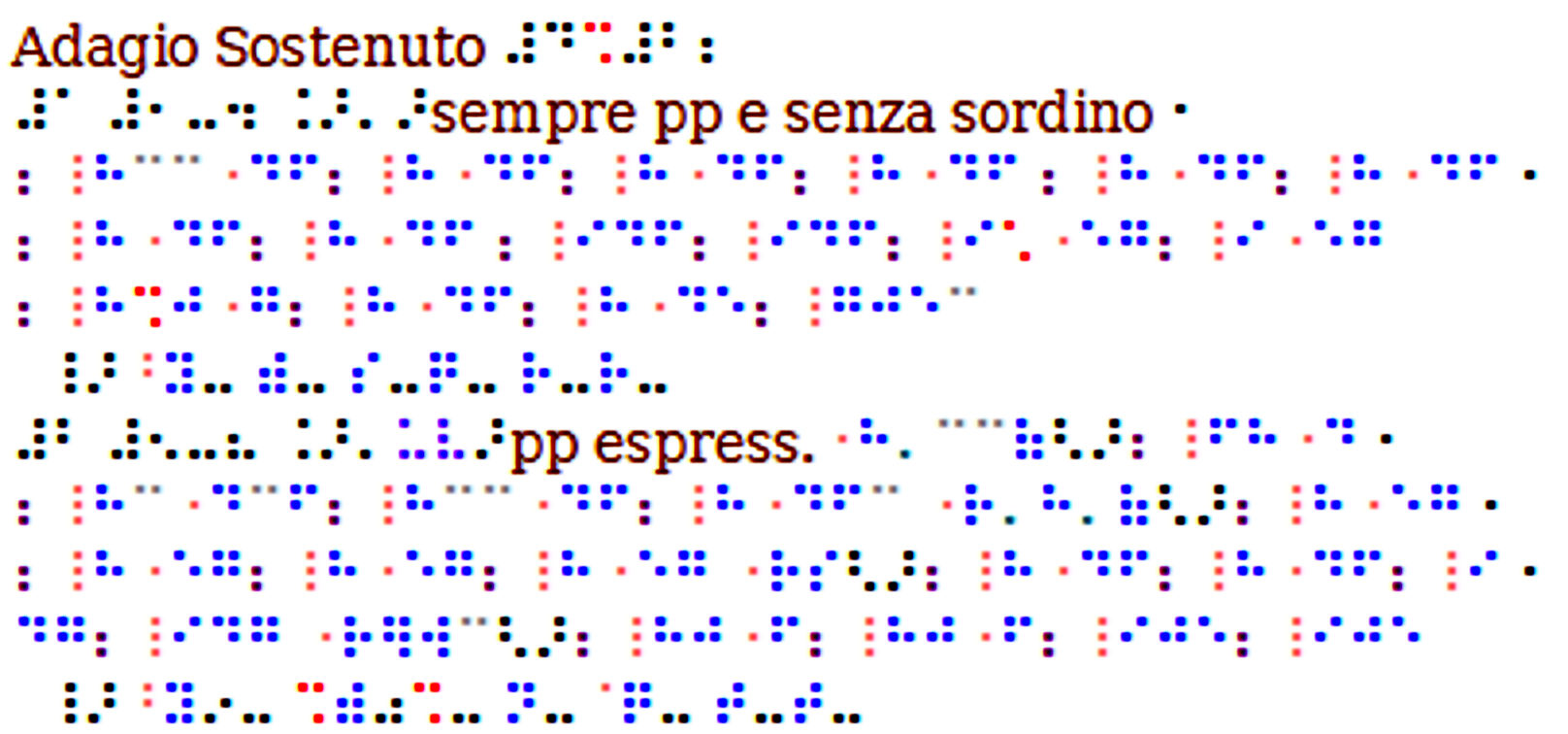} \\
     (c) & (d) \\
    \end{tabular}
\caption{Examples of scores written in various notations: (a) Common Western Music Notation (Dvorak Symphony No.9, IV), (b) White Mensural Notation (Belli \cite{Parada-Cabaleiro2017}), (c) Tabulature (Regondi, Etude No.10) and (d) Braille (Beethoven, Sonata No.14 Op.27 No.2).}
\label{fig:NotationExamples}
\end{figure}

Importantly, music can be written down in many different notation systems that can be seen as different languages to express musical concepts (see Fig. \ref{fig:NotationExamples}). \emph{CWMN} is probably the most prominent one. Before CWMN was established, other notations such as mensural or neumes preceded it, so we refer to them as \emph{historical notations}. Although this may seem like a tangential issue, the recognition of manuscripts in ancient notations has motivated a large number of works in OMR that facilitate the preservation and analysis of the cultural heritage as well as enabling digital musicological research of early music at scale \cite{Dalitz2008a, Vigliensoni2011, Fujinaga2014, Crawford2018}. Another category of notations that are still being actively used today are \emph{instrument-specific notations}, such as tablature for string instruments or percussion notation. The final category captures all \emph{other notations} including, e.g., modern graphic notation, braille music or numbered notation that are only rarely used and for which the existing body of music is much smaller than for the other notations.

To get an idea of how diverse the visual expression of music is, the Standard Music Font Layout \cite{Spreadbury2015} currently lists over 2440 recommended characters, plus several hundred optional glyphs.

Byrd and Simonsen \cite{Byrd2015} further characterize OMR inputs by the \emph{complexity} of the notated music itself, ranging from simple monophonic music to ``pianoform.'' They use both the presence of multiple staves as well as the number of notated voices inside a single staff as a dimension of notational complexity. In contrast, we do not see the number of staves as a driver of complexity since a page typically contains many staves and a decision on how to group them into systems has to be made anyway. Additionally, we explicitly add a category for \emph{homophonic} music that only has a single logical voice, even though that voice may contain chords with multiple notes being played simultaneously. The reason for singling out homophonic music is that once notes are grouped into chords, inferring onsets can dramatically be simplified into simply reading them left-to-right, as opposed to polyphonic music which requires the assignment of multiple logical voices first.

\begin{figure}
\centering
\begin{subfigure}[b]{\textwidth}
   \includegraphics[width=1\linewidth]{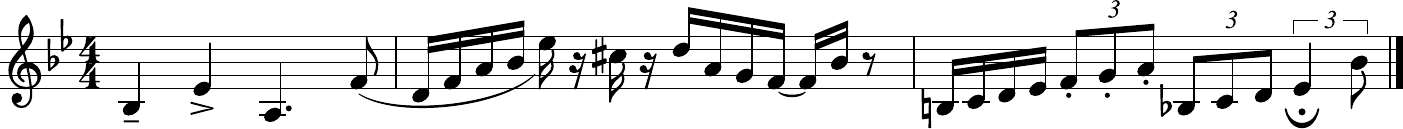}
   \caption{Monophonic}
   \label{fig:Monophonic}
\end{subfigure}
\begin{subfigure}[b]{\textwidth}
   \includegraphics[width=1\linewidth]{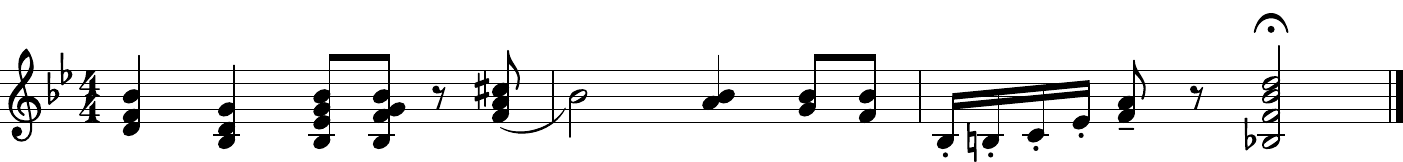}
   \caption{Homophonic}
   \label{fig:Homophonic}
   \end{subfigure}
\begin{subfigure}[b]{\textwidth}
   \includegraphics[width=1\linewidth]{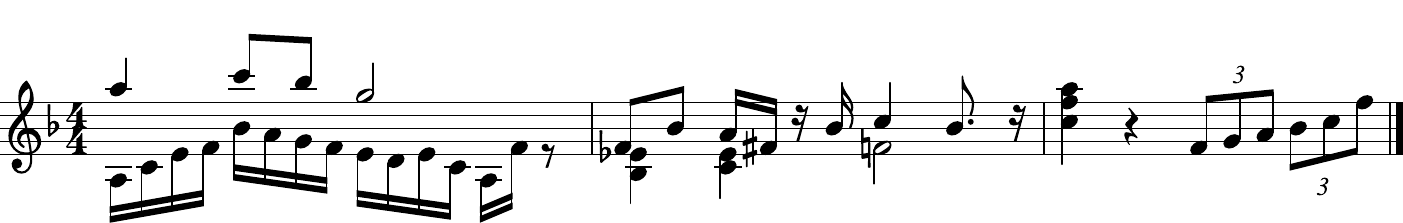}
   \caption{Polyphonic}
   \label{fig:Polyphonic}
\end{subfigure}
\begin{subfigure}[b]{\textwidth}
   \includegraphics[width=1\linewidth]{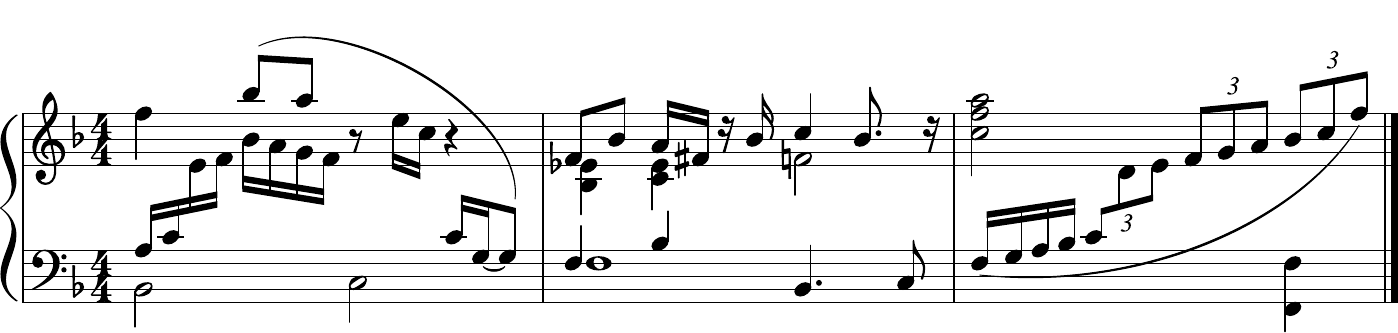}
   \caption{Pianoform}
   \label{fig:Pianoform}
\end{subfigure}
\caption{Examples of the four categories of structural complexity.}
\label{fig:StructuralComplexity}
\end{figure}

Therefore, we adapt Byrd and Simonsen's taxonomy \cite{Byrd2015} into the following four categories of \emph{structural complexity} (see Fig. \ref{fig:StructuralComplexity}):

\begin{enumerate}[label=(\alph*)]
    \item \emph{Monophonic}: only one note (per staff) is played at a time.
    \item \emph{Homophonic}: multiple notes can occur at the same time to build up a chord, but only as a single voice.
    \item \emph{Polyphonic}: multiple voices can appear in a single staff.
    \item \emph{Pianoform}: scores with multiple staves and multiple voices that exhibit significant structural interactions. They can be much more complex than polyphonic scores and cannot be disassembled into a series of monophonic scores, such as in polyphonic renaissance vocal part books.
\end{enumerate}

This complexity of the encoded music has significant implications on the whole OMR system because the various levels translate into different sets of constraints and requirements. Furthermore, the structural complexity cannot simply be adjusted or simulated like the visual complexity by applying an image operation on a perfect image \cite{Journet2017} because it represents an intrinsic property of the music.

Finally, as with other digital document processing, OMR inputs can be classified according to their image quality which is determined by two independent factors: the underlying \emph{document quality}, and the \emph{digital imaging acquisition} mode. The underlying document quality is a continuum on a scale from perfect or nearly flawless (e.g., if the document was born-digital and printed) to heavily degraded or defaced documents (e.g., ancient manuscripts that deteriorated over time and exhibit faded ink, ink blots, stains, or bleedthrough) \cite{Byrd2015}. The image acquisition mode is also a continuum that can extend from born-digital images, over scans of varying quality to low-quality, distorted photos that originate from camera-based scenarios with handheld cameras, such as smartphones \cite{Adamska2015, Vo2017}.

\subsection{OMR Outputs}
\label{sec:OmrOutputs}
The taxonomy of OMR \emph{outputs}, on the other hand, has not been treated as systematically in the OMR literature. Lists of potential or hypothetical applications are typically given in introductory sections \cite{Fujinaga1988, Blostein1992, Carter1992a, Novotny2015}. While this may not seem like a serious issue, it makes it hard to categorize different works and compare their results with each other because one often ends up comparing apples to oranges \cite{Bainbridge2001}.

The need for a more principled treatment is probably best illustrated by the unsatisfactory state of OMR evaluation. As pointed out by \cite{Byrd2015, Hajicjr.2016, Hajicjr.2018b}, there is still no good way at the moment of how to measure and compare the performance of OMR systems. The lack of such evaluation methods is best illustrated by the way OMR literature presents the state of the field: Some consider it a mature area that works well (at least for typeset music) \cite{Fornes2006, Fornes2008a, Baba2012, Baro2016, RicoBlanes2017}. Others describe their systems with reports of very high accuracies of up to nearly 100\% \cite{Sharif2009, Luangnapa2012, Wallner2014, Huang2015, Mehta2015, Nguyen2015, Pham2015, Vo2017, Calvo-Zaragoza2018}, giving an impression of success; however, many of these numbers are symbol detection scores on a small corpus with a limited vocabulary that are not straightforward to interpret in terms of actual usefulness, since they do not generalize \cite{Bellini2007, Byrd2015}.\footnote{The problem of results that cannot be compared was already noted in the very first review of OMR in 1972 by Kassler \cite{Kassler1972}, when he reviewed the first two OMR theses by Pruslin \cite{Pruslin1966} and Prerau \cite{Prerau1971}.} The existence of commercial applications \cite{SmartScore, PhotoScore, NotateMe, StaffPad, PlayScore, iSeeNotes, KompApp} is also sometimes used to support the claim that OMR ``works'' \cite{Baro2018}. On the other hand, many researchers think otherwise \cite{Byrd2006, Bellini2007, Rebelo2012, Ng2014, Chen2016, Choi2017, Pacha2017, Riba2017, Hajicjr.2018, Hajicjr.2018a}, emphasizing that OMR does not provide satisfactory solutions in general---not even for typeset music. Some indirect evidence of this can be gleaned from the fact that even for high-quality scans of typeset music, only a few projects rely on OMR,\footnote{Some users of the Choral Public Domain Library (CPDL) project use commercial applications such as SharpEye or PhotoScore Ultimate: \url{http://forums.cpdl.org/phpBB3/viewtopic.php?f=9&t=9392}, but without recording the time
spent in correcting and editing the output.} while other projects still prefer to
crowdsource manual transcriptions instead of using systems for automatic recognition \cite{Gotham2018}, or at least crowdsource the correction of the errors produced by OMR systems \cite{Saitis2014}. Given the long-standing absence of OMR evaluation standards, this ambivalence is not surprising. However, a scientific field should be able to communicate its results in comprehensible terms to external stakeholders---something OMR is currently unable to do.

We feel that to a great extent this confusion stems from the fact that the question ``Does OMR work?'' is an overly vague question. As our analysis in Section \ref{sec:WhatIsOpticalMusicRecognition} shows, OMR is not a monolithic problem---therefore, asking about the ``state of OMR'' is \emph{under-specified}. ``Does OMR work?'' must be followed by ``... as a tool for X,'' where X is some application, in order for such questions to be answerable. There is, again, evidence for this in the OMR literature. OMR systems have been properly evaluated in retrieval scenarios \cite{Balke2015, Fremerey2008, Achankunju2018} or in the context of digitally replicating a musicological study \cite{Hajicjr.2018}. It has, in fact, been explicitly asserted \cite{Hajicjr.2018b} that evaluation methodologies are only missing for a limited subset of OMR applications.
In particular, there is no known meaningful edit distance between two scores, i.e., an algorithm that computes the similarity between two scores and produces the steps that are needed to transform one score into the other. If such a distance function existed, one could at least count the number of required edits to assess the performance of an OMR system quantitatively. However, this does not necessarily provide a good measure of quality because it is unclear how to weight the costs of different edit operations, e.g., getting a note duration wrong vs. missing an articulation mark.

At the same time, the granularity at which we define the various tasks should not be too fine, otherwise one risks entering a different swamp: instead of no evaluation at all, each individual work is evaluated on the merits of a narrowly defined (and often merely hypothetical) application scenario, which also leads to contributions that cannot be compared. In fact, this risk has already been illustrated on the subtask of symbol detection, which seems like a well-defined problem where the comparison should be trivial. In 2018, multiple music notation object detection papers were published \cite{Pacha2018, Pacha2018b, Hajicjr.2018a, Tuggener2018}, but each reported results in a different way while presenting a good argument for choosing that kind of evaluation, so significant effort was necessary in order to make these contributions directly comparable \cite{Pacha2018c}. A compromise is therefore necessary between fully specifying the question of whether OMR ``works'' by asking for a specific application scenario, and on the other hand retaining sufficiently general categories of such tasks.

Having put forward the reasoning why systematizing the field of OMR with respect to its outputs is desirable, we proceed to do so. For defining meaningful categories of outputs for OMR, we come back to the fundamentals of how OMR inverts the music encoding process to recover the musical semantics and musical notation (see Section \ref{sec:WhatIsOpticalMusicRecognition}). These two prongs of reading musical documents roughly correspond to two broad areas of OMR applications \cite{Miyao2000} that overlap to a certain extent:

\begin{itemize}
\item \emph{Replayability}: recovering the encoded music itself in terms of pitch, velocity, onset, and duration. This application area sees OMR as a component inside a bigger music processing pipeline that enables the system to operate on music notation documents as just another input. Notice that readability by humans is not required for these applications, as long as the computer can process and ``play'' the symbolic data.
\item \emph{Structured Encoding}: recovering the music along with the information on how it was encoded using elements of music notation. This avenue is oriented towards providing the score for music performance, which requires a (lossless) re-encoding of the score, assuming that humans read the OMR output directly. Recovering the musical semantics might not in fact be strictly necessary, but in practice, one often wishes to obtain that information too, in order to enable digitally manipulating the music in a way that would be easiest done with the semantics being recovered (e.g., transposing a part to make it suitable for another instrument).
\end{itemize}

In other words, the output of an application that targets replayability is typically processed by a machine, whereas humans usually demand the complete recognition of the structured encoding to allow for a readable output (see Fig. \ref{fig:SchumannWithGoodAndBadEngraving}).

\begin{figure}[ht]
\includegraphics[width=\textwidth]{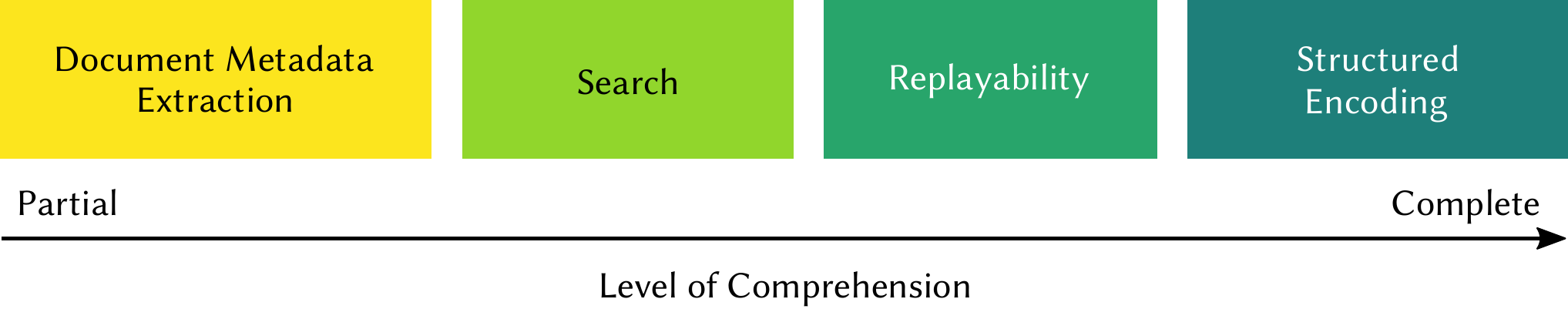}
\caption{Taxonomy of four categories of OMR applications that require an increasing level of comprehension, starting with metadata extraction where a minimal understanding might be sufficient, up to structured encoding that requires a complete understanding of music notation with all its intricacies.}
\label{fig:LevelsOfComprehension}
\end{figure}

While the distinction between replayability and structured encoding is already useful, there are other reasons that make it interesting to read musical notation from a document. For example, to search for specific content or to draw paleographic conclusions about the document itself. Therefore, we need to broaden the scope of OMR to actually capture these applications. We realized that some use-cases require much less comprehension of the input and music notation than others. To account for this, we propose the following four categories that demand an increasing level of comprehension: \emph{Document Metadata Extraction, Search, Replayability}, and \emph{Structured Encoding} (see Fig. \ref{fig:LevelsOfComprehension}).

Depending on the goal, applications differ quite drastically in terms of requirements---foremost in the choice of output representation. Furthermore, this taxonomy allows us to use different evaluation strategies.

\subsubsection{Document Metadata Extraction}
\label{sec:DocumentMetadataExtraction}
The first application area requires only a partial understanding of the entire document and attempts to answer specific questions about it. These can be very primitive ones, like whether a document contains music scores or not, but the questions can also be more elaborate, for example:

\begin{itemize}
    \item In which period was the piece written?
    \item What notation was used?
    \item How many instruments are involved?
    \item Are two segments written by the same copyist?
\end{itemize}

All of the aforementioned tasks entail a different level of underlying computational complexity. However, we are not organizing applications according to their difficulty but instead by the type of answer they provide. In that sense, all of these tasks can be formulated as classification or regression problems, for which the output is either a discrete category or a continuous value, respectively.

\begin{definition} \label{def:DocumentMetadataExtraction}
Document metadata extraction refers to a class of Optical Music Recognition applications that answer questions about the music notation document.
\end{definition}

The output representation for document metadata extraction tasks are scalar values or category labels.
Again, this does not imply that extracting the target values is necessarily easy, but that the difficulties are not related to the output representation, as is the case for other uses.

Although this type of application has not been very popular in the OMR literature, there is some work that approaches this scenario. In \cite{Bainbridge2006} and \cite{Pacha2017} the authors describe systems that classify images according to whether they depict music scores or not. While the former one used a basic computer vision approach with a Hough transform and run-length ratios, the latter uses a deep convolutional neural network. Such systems can come in handy if one has to automatically classify a very large number of documents \cite{Paeaekkoenen2018}. Perhaps the most prominent application is identifying the writer of a document \cite{Fornes2008, Fornes2009, Gordo2013, Roy2017} (which can be different from the composer). This task was one of the main motivations behind the construction of the CVC-MUSCIMA dataset \cite{Fornes2012} and was featured in the ICDAR 2011 Music Score Competition \cite{Fornes2011}.

The document metadata extraction scenario has the advantage of its unequivocal evaluation protocols. Tasks are formulated as either classification or regression problems, and these have well-defined metrics such as accuracy, F-measure, or mean squared error.

\subsubsection{Search}
\label{sec:Search}
Nowadays we have access to millions of musical documents. Libraries and communities have taken considerable efforts to catalog and digitize music scores, by scanning them and freely providing users access to them, e.g., IMSLP \cite{IMSLP}, SLUB \cite{SLUB}, DIAMM \cite{DIAMM} or CPDL \cite{CPDL}, to name a few. Here is a fast growing interest in automated methods which would allow users to search for relevant musical content inside these sources systematically. Unfortunately, searching for specific content often remains elusive because many projects only provide the images and manually entered metadata. We capture all applications that enable such lookups under the category \emph{Search}. Examples of search questions could be:

\begin{itemize}
\item Do I have this piece of music in my library?
\item On which page can I find this melody?
\item Where does this sequence of notes (e.g., a theme) repeat itself?
\item Was a melody copied from another composition?
\item Find the same measure in different editions for comparing them.
\end{itemize}

\begin{definition} \label{def:Search}
Search refers to a class of Optical Music Recognition applications that, given a collection of sheet music and a musical query, compute the relevance of individual items of the collection with respect to the given query.
\end{definition}

Applications from this class share a direct analogy with keyword spotting (KWS) in the text domain \cite{Giotis2017} and a common formulation: the input is a query as well as the collection of documents on which the search is to be carried out; the output is the selection of elements from that collection that match the query. However, that selection of elements is a loose concept and can refer to a complete music piece, a page, or in the most specific cases, a particular bounding-box or even a pixel-level location. In the context of OMR, the musical query must convey musical semantics (as opposed to general search queries, e.g., by title or composer; hence the term ``musical'' query in Definition \ref{def:Search}). The musical query is typically represented in a symbolic way, interpretable unambiguously by the computer (similar to query-by-string in KWS), yet it is also interesting to consider queries that involve other modalities, such as image queries (query-by-example in KWS) or audio queries (query-by-humming in audio information retrieval or query-by-speech in KWS). Additionally, it makes sense to establish different domain-specific types of matching, as it is useful to perform searches restricted to specific music concepts such as melodies, sequences of intervals, or contours, in addition to exact matching.

A direct approach for search within music collections is to use OMR technology to transform the documents into symbolic pieces of information, over which classical content-based or symbolic retrieval methods can be used \cite{Choudhury2000, Barton2002, Dovey2004, Thompson2011, Helsen2014, Keil2017, Achankunju2018, Diet2018a}. The problem is that these transformations require a more comprehensive understanding of the processed documents (see Sections \ref{sec:Replayability} and \ref{sec:StructuredEncoding} below). To avoid the need for an accurate symbol-by-symbol transcription, search applications can resort to other methods to determine whether (or how likely) a given query is in a document or not. For instance, in cross-modal settings, where one searches a database of sheet music using a MIDI file \cite{Fremerey2008, Balke2015} or a melodic fragment that is given by the user on the fly \cite{Achankunju2018}, OMR can be used as a hash function. When the queries and documents are both projected into the search space by the same OMR system, some limitations of the system may even cancel out (e.g., ignoring key signatures), so that retrieval performance might deteriorate less than one would expect. Unfortunately, if either the query or the database contains the true musical semantics, such errors do become critical \cite{Hajicjr.2018}.

A few more works have also approached the direct search of music content without the need to convert the documents into a symbolic format first. Examples comprise the works by Malik et al. \cite{Malik2013} dealing with a query-by-example task in the CVC-MUSCIMA dataset, and by Calvo-Zaragoza et al. \cite{Calvo-Zaragoza2018e}, considering a classical query-by-string formulation over early handwritten scores. In the cross-modal setting, the audio-sheet music retrieval contributions of Dorfer et al. \cite{Dorfer2018} are an example of a system that explicitly attempts to gain only the minimum level of comprehension of music notation necessary for performing its retrieval job.

Search systems usually retrieve not just a single result but all those that match the input query, typically sorted by confidence. This setting can re-use general information retrieval methodologies for evaluating performance, such as precision and recall as well as encompassing metrics like average precision and mean average precision \cite{Manning2008, Harman2011}.

\subsubsection{Replayability}
\label{sec:Replayability}
Replayability applications are concerned with reconstructing the notes encoded in the music notation document. Notice that producing an actual audio file is not considered to be part of OMR, despite being one of the most frequent use-cases of OMR. In any case, OMR can enable these applications by recovering the pitches, velocities, onsets, and durations of notes. This symbolic representation, usually stored as a MIDI file, is already a very useful abstraction of the music itself and allows for plugging in a vast range of computational tools such as:
\begin{itemize}
\item synthesis software to produce an audio representation of the composition
\item music information retrieval tools that operate on symbolic data
\item tools that perform large-scale music-theoretical analysis
\item creativity-focused applications \cite{Xia2017}
\end{itemize}

\begin{definition} \label{def:Replayability}
Replayability refers to a class of Optical Music Recognition applications that recover sufficient information to create an audible version of the written music.
\end{definition}

Producing a MIDI (or an equivalent) representation is one key goal for OMR---at least for the foreseeable future since MIDI is a representation of music that has a long tradition of computational processing for a vast variety of purposes. Many applications have been envisioned that only require replayability. For example, applications that can sight-read the scores to assist practicing musicians by providing missing accompaniment.

Replayability is also a major concern for digital musicology. Historically, the majority of compositions have probably never been recorded, and therefore are only available in written form as scores; of these, most compositions have also never been typeset, since typesetting has been a very expensive endeavor, reserved essentially either for works with assured commercial success, or composers with substantial backing by wealthy patrons. Given the price of manual transcription, it is prohibitive to transcribe large historical archives. At the present time, OMR that produces MIDI, especially if it can do so for manuscripts, is probably the only tool that could open up the vast amount of compositions to quantitative musicological research, which, in turn, could perhaps finally begin to answer broad questions about the evolutions of the average musical styles, instead of just relying on the works of the relatively few well-known composers.

Systems designed for the goal of replayability traditionally seek first to obtain the structured encoding of the score (see Section \ref{sec:StructuredEncoding}), from which the sequences of notes can be straightforwardly retrieved \cite{Hajicjr.2018a}. However, if the specific goal is to obtain something equivalent to a MIDI representation, it is possible to simplify the recognition and ignore many of the elements of musical notation, as demonstrated by numerous research projects \cite{Matsushima1985, Baumann1992, Homenda1996, Fotinea2000, Rossant2004, Huang2015, Pacha2018b}. An even clearer example of this distinction can be observed in the works of Shi et al. \cite{Shi2017} as well as van der Wel and Ullrich \cite{Wel2017}; both focus only on obtaining the sequence of note pairs (duration, pitch) that are depicted in single-staff images, regardless of how these notes were actually expressed in the document. Another instance of a replay-oriented application is the Gocen system \cite{Baba2012} that reads handwritten notes with a specially designed device with the goal of producing a musical performance while ignoring the majority of music notation syntax.

Once a system is able to arrive at a MIDI-like representation, evaluating the results is a matter of comparing sequences of pitch-onset-duration-triplets. Velocities may optionally be compared too, once the note-by-note correspondence has been established, but can be seen as secondary for many applications. Note, however, that even on the level of describing music as configurations of pitch-velocity-onset-duration-quadruples, MIDI is a further simplification that is heavily influenced by its origin as a digital representation of performance, rather than a composition: the most obvious inadequacy of MIDI is its inability to distinguish pitches that sound equivalent but are named differently, e.g., F-sharp and G-flat.\footnote{This is the combined heritage of equal temperament, where these two pitches do correspond to the same fundamental frequency, and of the origins of MIDI in genres dominated by fretted and keyboard instruments.}

Multiple similarity metrics for comparing MIDI files have been proposed during the Symbolic Melodic Similarity track of the Music Information Retrieval Evaluation eXchange (MIREX),\footnote{ \url{https://www.music-ir.org/mirex/wiki/MIREX_HOME}} e.g., by determining the local alignment between the geometric representations of the melodies \cite{Urbano2010, Urbano2011, Urbano2012, Urbano2013}. Other options could be multi-pitch estimation evaluation metrics \cite{Bay2009}, Dynamic Time Warping \cite{Dorfer2018}, or edit distances between two time-ordered sequences of pitch-duration pairs \cite{Zhang2017, Calvo-Zaragoza2018}.

\subsubsection{Structured Encoding}
\label{sec:StructuredEncoding}
It can be reasonably stated that digitizing music scores for ``human consumption'' and score manipulation tasks that a \emph{vollkommener Capellmeister}\footnote{roughly translated from German as ``ideal conductor''} \cite{Mattheson1739} routinely performs, such as part exporting, merging, or transposing for available instruments is the original motivation of OMR ever since it started \cite{Pruslin1966, Prerau1971, Fujinaga1988, Bainbridge1997} and the one that appeals to the widest audience. Given that typesetting music is troublesome and time-consuming, OMR technology represents an attractive alternative to obtain a digital version of music scores on which these operations can be performed efficiently with the assistance of the computer.

This brings us to our last category that requires the highest level of comprehension, called structured encoding. Structured encoding aims to recognize the entire music score while retaining all the engraving information available to a human reader. Since there is no viable alternative to music notation, the system has to fully transcribe the document into a structured digital format with the ultimate goal of keeping the same musical information that could be retrieved from the physical score itself.

\begin{definition} \label{def:StructuredEncoding}
Structured Encoding refers to a class of Optical Music Recognition applications that fully decode the musical content, along with the information of 'how' it was encoded by means of music notation.
\end{definition}

\begin{figure}
\includegraphics[width=\textwidth]{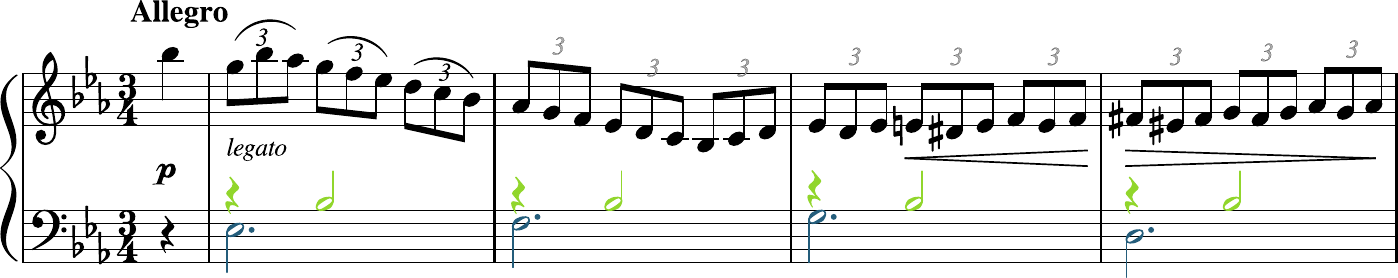}
\caption{Beginning of Franz Schubert, Impromptu D.899 No. 2. The triplet marks starting in the second measure of the top staff are typically omitted in printed editions (here depicted in gray for visualization). The two distinct voices in the bottom staff are color-coded in green and blue.}
\label{fig:SchubertImpromptuNo2}
\end{figure}

Note that the difference between replayability and structured encoding can seem vague: for instance, imagine a system that detects all notes and all other symbols and exports them into a MusicXML file. The result, however, is not the structured encoding unless the system also attempts to preserve the information on how the scores were laid out. That does not mean it has to store the bounding box and exact location of every single symbol, but the engraving information that \emph{conveys musical semantics}, like whether the stem of a note went up or down. To illustrate this, consider the following musical snippet in Fig. \ref{fig:SchubertImpromptuNo2}. If a system like the one described in \cite{Calvo-Zaragoza2018} recognized this, it would remain restricted to replayability. Not because of the current limitations to monophonic, single-staff music, but due to the selected output representation, which does not store engraving information such as the simplifications that start in the second measure of the top staff (the grayed out 3s that would be omitted in the printing) or the stem directions of the notes in the bottom staff (green and blue) that depict two different voices. In summary, any system discarding engraving information that conveys musical semantics, by definition, cannot reach the structured encoding goal.

To help understand why structured encoding poses such a difficult challenge, we would like to avail ourselves of the intuitive comparison given by Donald Byrd:\footnote{\url{http://music.informatics.indiana.edu/don_notation.html}} representing music as time-stamped events (e.g., with MIDI) is similar to storing a piece of writing in a plain text file; whereas representing music with music notation (e.g., with MusicXML) is similar to a structured description like an HTML website. By analogy, obtaining the structured encoding from the image of a music score can be as challenging as recovering the HTML source code from the screenshot of a website.

Since this use-case appeals to the widest audience, it has seen development both from the scientific research community and commercial vendors. Notable products that attempt full structured encoding include SmartScore \cite{SmartScore}, Capella Scan \cite{CapellaScan}, PhotoScore \cite{PhotoScore} as well as the open-source application Audiveris \cite{Audiveris}. While the projects described in many scientific publications seem to be striving for structured encoding to enable interesting applications such as the preservation of the cultural heritage \cite{Chen2014}, music renotation \cite{Chen2015}, or transcriptions between different music notation languages \cite{Rizo2018}, we are not aware of any systems in academia that would actually produce structured encoding.

A major stumbling block for structured encoding applications has for a long time been the lack of practical formats for representing music notation that would be powerful enough to retain the information from the input score, and at the same time be a natural endpoint for OMR. This is illustrated by papers that propose OMR-specific representations, both before the emergence of MusicXML \cite{Good2001, Good2003} as a viable interchange format \cite{Miyao2000} and after \cite{Hajicjr.2017d}. At the same time, however, even without regard for OMR, there are ongoing efforts to improve music notation file formats: further development of MusicXML has moved into the W3C Music Notation Community Group,\footnote{\url{https://www.w3.org/community/music-notation/}} and there is an ongoing effort in the development of the Music Encoding Initiative format \cite{Roland2002}, best illustrated by the annual Music Encoding Conference.\footnote{\url{https://music-encoding.org/conference/past.html}} Supporting the whole spectrum of music notation situations that arise in a reasonably-sized archive is already a difficult task. This can be evidenced by the extensive catalog of requirements for music notation formats that Byrd and Isaacson \cite{Byrd2016} list for a multi-purpose digital archive of music scores. Incidentally, the same paper also mentions support for syntactically incorrect scores among the requirements, which is one of the major problems that OMR has with outputting to existing formats directly. Although these formats are becoming more precise and descriptive, they are not designed to store information about how the content was automatically recognized from the document. This kind of information is actually relevant for evaluation, as it allows, for example, determining if a pitch was misclassified because of either a wrongly detected position in the staff or a wrongly detected clef.

The imperfections of representation standards for music notation is also reflected in a lack of evaluation standards for structured encoding. Given the ground truth representation of a score and the output of a recognition system, there is currently no automatic method that is capable of reliably computing how well the recognition system performed. Ideally, such a method would be rigorously described and evaluated, have a public implementation, and give meaningful results. Within the traditional OMR pipeline, the partial steps (such as symbol detection) can use rather general evaluation metrics. However, when OMR is applied for getting the structured encoding of the score, no evaluation metric is available, or at least generally accepted, partially because of the lack of a standard representation for OMR output, as mentioned earlier. The notion of ``edit cost'' or ``recognition gain'' that defines success in terms of how much time a human editor saves by using an OMR system is yet more problematic, as it depends on the editor and on the specific toolchain \cite{Bellini2007}.

Haji{\v{c}} \cite{Hajicjr.2018b} argues that a good OMR evaluation metric should be intrinsic\footnote{Extrinsic evaluation means evaluating the system in an application context: ``How good is this system for application X?.'' Intrinsic evaluation attempts to evaluate a system without reference to a specific use-case, asking how much of the encoded information has been recovered. In the case of OMR, this essentially reduces evaluation to error counting.} and independent of a certain use-case. The benefits would be the independence from the selected score editing toolchain as well as the music notation format and a clearly interpretable automatic metric for guiding OMR development (which could ideally be used as a differentiable loss function for training full-pipeline end-to-end machine learning-based systems).

There is no reason why a proper evaluation should not be possible since there is only a finite amount of information that a music document retains, which can be exhaustively enumerated. It follows that we should be able to measure what proportion of this information our systems recovered correctly. One reason why this is still such a hard problem is the lack of an underlying formal model of music notation for OMR. Such a model could support structured encoding evaluation by being:

\begin{itemize}
\item \emph{Comprehensive}: integrating naturally both the ``reprintability'' and ``replayability'' level (also called graphical and semantic level in the literature), by being capable of describing the various corner cases (which implies extensibility);
\item \emph{Useful}: enabling tractable inference (at least approximate) and an adequate distance function; and
\item \emph{Sufficiently supported} through open-source software.
\end{itemize}

Historically, context-free grammars have been the most explored avenue for a unified formal description of music notation, both with an explicit grammar \cite{Andronico1982, Coueasnon1994} and implicitly using a modified stack automaton \cite{Bainbridge2003}: this feels natural, given that music notation has strict syntactic rules and hierarchical structures that invite such descriptions. The 2-D nature of music notation also inspired graph grammars \cite{Fahmy1993a} and attributed graph grammars \cite{Baumann1995}. Recently, modeling music notation as a directed acyclic graph has been proposed as an alternative \cite{Hajicjr.2017d, Hajicjr.2018a}. However, none of these formalisms has yet been adopted: the notation graph is too recent and does not have sufficient software and community support, and the older grammar-based approaches lack up-to-date open-source implementations altogether (and are insufficiently detailed in the respective publications for re-implementation). Without an appropriate formalism and the corresponding tooling, the evaluation of structured encoding can hardly hope to move beyond ad-hoc methods.

It is important to distinguish between the formal model of music notation on which OMR systems operate, and the output representation they generate. MusicXML and MEI are good choices as output format because of their widespread tool support and acceptance, but are not suitable as internal representation of OMR systems. For example, they describe music scores top-down, something like "page-staff-measure-note" with semantic terms such as notes that have a pitch and a duration whereas an OMR system typically detects music bottom-up, often starting with abstract graphical shapes at certain positions in the image that need to be stored and processed somehow before the actual semantics can be inferred. MusicXML and MEI were simply not designed to capture these low-level graphical concepts but the final, well-defined musical concepts. That is why almost all OMR systems have their own internal model before encoding the results into standard formats.

\section{Approaches to OMR}
\label{sec:ApproachesToOmr}
In order to complete our journey through the landscape of \emph{Optical Music Recognition}, we yet have to visit the arena of OMR techniques. These have recently seen a strong shift towards machine learning that has brought about a need to revisit the way that OMR methods have traditionally been systematized. As opposed to OMR applications, the vocabulary of OMR methods and subtasks already exists \cite{Rebelo2012} and only needs to be updated to reflect the new reality of the field.

As mentioned before, obtaining the structured encoding of the scores has been the main motivation for developing the OMR field. Given the difficulty of such objective, the process was usually approached by dividing it into smaller stages that could represent challenges within reach with the available technologies and resources. Over the years, the pipeline described by Bainbridge and Bell \cite{Bainbridge2001}, refined by Rebelo et al. in 2012 \cite{Rebelo2012} became the de-facto standard. That pipeline is traditionally organized into the following four blocks, sometimes with slightly varying names and scopes of the individual stages:

\begin{enumerate}
\item \emph{Preprocessing}: Standard techniques to ease further steps, e.g., contrast enhancement, binarization, skew-correction or noise removal. Additionally, the layout should be analyzed to allow subsequent steps to focus on actual content and ignore the background.
\item \emph{Music Object Detection}: Finding and classifying all relevant symbols or glyphs in the image.
\item \emph{Notation Assembly}: Recovering the music notation semantics from the detected and classified symbols. The output is a symbolic representation of the symbols and their relationships, typically as a graph.
\item \emph{Encoding}: Encoding the music into any output format unambiguously, e.g., into MIDI for playback or MusicXML/MEI for further editing in a music notation program.
\end{enumerate}

With the appearance of deep learning in OMR, many steps that traditionally produced suboptimal results, such as the staff-line removal or symbol classification have seen drastic improvements \cite{Gallego2017, Pacha2017} and are nowadays considered solved or at least clearly solvable. This caused some steps to become obsolete or collapse into a single (bigger) stage. For instance, the music object detection stage was traditionally separated into a segmentation stage and a classification stage. Since staff lines make it hard to separate isolated symbols through connected component analysis, they typically were removed first, using a separate method. However, deep learning models with convolutional neural networks have been shown to be able to deal with the music object detection stage holistically without having to remove staff lines at all. In addition to the performance gains, a compelling advantage is the capability of these models to be trained in a single step by merely providing pairs of images and positions of the music objects to be found, eliminating the preprocessing step altogether. A baseline of competing approaches on several datasets containing both handwritten and typeset music can be found in the work of Pacha et al. \cite{Pacha2018c}.

The recent advances also diversified the way OMR is approached altogether: there are alternative pipelines with their own ongoing research that attempt to face the whole process in a single step. This holistic paradigm, also referred to as end-to-end systems, has been dominating the current state of the art in other tasks such as text, speech, or mathematical formula recognition \cite{Chowdhury2018, Chiu2018, Zhang2017}. However, due to the complexity of inferring musical semantics from the image, it is difficult (for now) to formulate it as a learnable optimization problem. While end-to-end systems for OMR do exist, they are still limited to a subset of music notation, at best. Pugin pioneered this approach utilizing hidden Markov models for the recognition of typeset mensural notation \cite{Pugin2006}, and some recent works have considered deep recurrent neural networks for monophonic music written in both typeset \cite{Shi2017, Wel2017, Calvo-Zaragoza2018, Calvo-Zaragoza2018b} and handwritten \cite{Baro2018} modern notation. However, polyphonic and pianoform scores are currently out of reach because there is simply no appropriate model formulation. Another limitation of these approaches is that they do not allow inspecting intermediate results to identify errors.

Along with the shift towards machine learning, several public datasets have emerged, such as MUSCIMA++ \cite{Hajicjr.2017d}, DeepScores \cite{Tuggener2018} or Camera-PrIMuS \cite{Calvo-Zaragoza2018b}.\footnote{A full list of all available datasets can be found at \url{https://apacha.github.io/OMR-Datasets/}} There are also significant efforts to develop tools by which training data for OMR systems can be obtained including MUSCIMarker \cite{Hajicjr.2017c}, Pixel.js \cite{Saleh2017}, and MuRET \cite{Rizo2018}.
On the other hand, while the machine learning paradigm has undeniably brought significant progress, it has shifted the costs onto data acquisition. This means that while machine learning can be more general and deliver state-of-the-art results when appropriate data is available, it does not necessarily drive down the costs of applying OMR. Still, we would say---tentatively---that once these resources are spent, the chances of OMR yielding useful results for the specific use-case are higher compared to earlier paradigms.

Tangentially to the way of dealing with the process itself, there has been continuous research on interactive systems for years. The idea behind such systems is based on the insight that OMR systems might always make some errors, and if no errors can be tolerated, the user is essential to correct the output. These systems attempt to incorporate user feedback into the OMR process in a more efficient way than just post-processing system output. Most notably is the interactive system developed by Chen et al. \cite{Chen2017, Chen2018}, where users directly interact with the OMR system by specifying which constraints to take into account while visually recognizing the scores. Users can then iteratively add or remove constraints before re-recognizing individual measures until they are satisfied. The most powerful feature of interactive systems is probably the displaying of recognition results superimposed on top of the original image, allowing errors to be spotted quickly \cite{CapellaScan, Audiveris, Vigliensoni2018, Rizo2018}.

\section{Conclusions}
\label{sec:Conclusion}
In this article, we have first addressed what \emph{Optical Music Recognition} is and proposed to define it as research field that investigates how to computationally read music notation in documents---a definition that should adequately delimit the field, and set it in relation to other fields such as OCR, graphics recognition, computer vision, and fields that await OMR results. We furthermore analyzed in depth the inverse relation of OMR to the process of writing down a musical composition and highlighted the relevance of engraving music properly---something that must also be recognized to ensure readability for humans. The investigation of what OMR is, revealed why this seemingly easy task of reading music notation has turned out to be such a hard problem: besides the technical difficulties associated with document analysis, many fundamental challenges arise from the way how music is expressed and captured in music notation. By providing a sound, concise, and inclusive definition, we capture how the field sees and talks about itself.

\begin{figure}
\includegraphics[width=\textwidth]{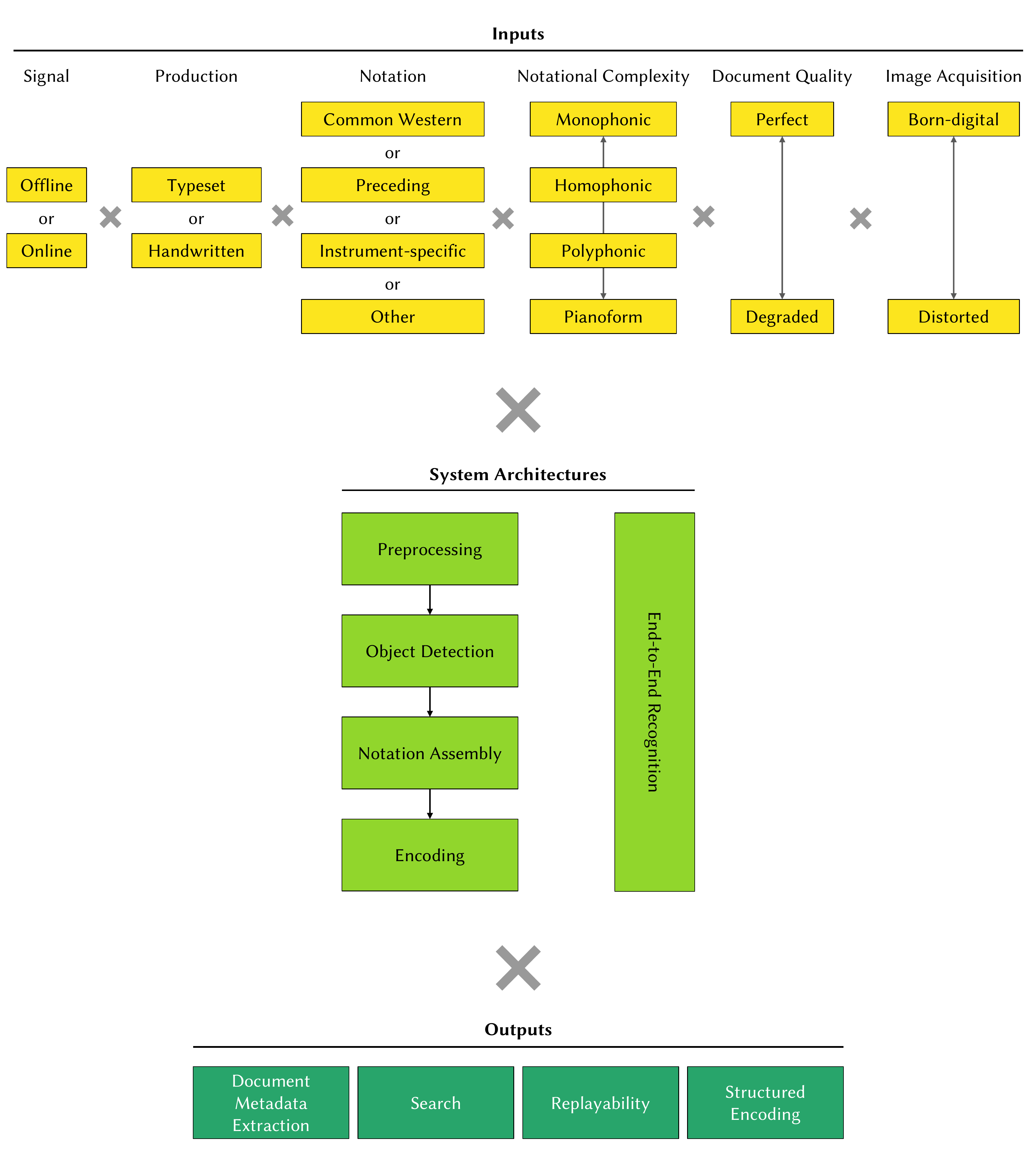}
\caption{An overview of the taxonomy of OMR inputs, architectures, and outputs. A fairly simple OMR system could, for example, read high-quality scans (offline) of well-preserved documents that contain typeset, monophonic, mensural notation, process it in a traditional, modular pipeline and write the results into a MIDI file to achieve replayability. A more sophisticated system, on the other hand, would allow images (offline) of handwritten music in common western notation from degraded documents as input and strive to recognize the full structured encoding in an end-to-end fashion. Note that the distinction between modular systems and end-to-end recognition is not binary as there can be hybrid systems that only merge a subset of the stages.}
\label{fig:TaxonomyOverview}
\end{figure}

We have then reviewed and improved the taxonomy of OMR, which should help systematize the current and future contributions to the field. While the inputs of OMR systems have been described systematically and established throughout the field, a taxonomy of OMR outputs and applications has not been proposed before. An overview of this taxonomy is given in Fig. \ref{fig:TaxonomyOverview}.

Finally, we have also updated the general breakdown of OMR systems into separate subtasks in order to reflect the recent advances with machine learning methods and discussed alternative approaches such as end-to-end systems and interactive scenarios.

One of the key points we wanted to stress is the internal diversity of the field: OMR is not a monolithic task. As analyzed in Section \ref{sec:InvertingTheMusicEncodingProcess}, it enables various use-cases that require fundamentally different system designs, as discussed in Section \ref{sec:OmrOutputs}. So before creating an OMR system, one should be clear about the goals and the associated challenges.

The sensitivity to errors is another relevant issue that needs to be taken into account. As long as errors are inevitable \cite{Chen2018, Crawford2018}, it is important to consider the impact of those errors to the envisioned application. If someone wants to transcribe a score with an OMR system, but the effort needed for correcting the errors is greater than the effort for directly entering the notes into a music notation program, such an OMR system would obviously be useless \cite{Bellini2007}. Existing literature on error-tolerance is inconclusive: while we tend to believe that users---especially practicing musicians---would not tolerate false recognitions \cite{Roggenkemper2018}, we also see systems that can handle a substantial amount of errors \cite{Crawford2018, Achankunju2018, Hajicjr.2018} and still produce meaningful results, e.g., when searching in a large database of scores. Therefore, it cannot be decided in advance how severe errors are, as it is always the end user who sets the extent of tolerable errors.

Readers should now comprehend the spectrum of what OMR might do, understand the challenges that reading music notation entails, and have a solid basis for further exploring the field on their own---in other words, be equipped to address the issues described in the next section.

\subsection{Open Issues and Perspectives for Future Research}
\label{sec:OpenIssuesAndPerspectivesForFutureResearch}
We conclude this paper by listing major open problems in Optical Music Recognition that significantly impede its progress and usefulness. While some of them are technical challenges, there are also many non-technical issues:

\begin{itemize}
\item \emph{Legal aspects}: Written music is the intellectual property of the composer and its allowed uses are defined by the respective publisher. Recognizing and sharing music scores can be seen as copyright infringement, like digitizing books without permission. To avoid this dispute, many databases such as IMSLP only store music scores whose copyright protection has expired. So an OMR dataset is either limited to old scores or one enters a legal gray area if not paying close attention to the respective license of every piece stored therein.
\item \emph{Stable community}: For decades, OMR research was conducted by just a few individuals whose work was distributed and mostly uncoordinated. Most OMR researchers joined the field with minor contributions but left again soon afterward. Furthermore, due to a lack of dedicated venues, researchers rarely met in person \cite{Calvo-Zaragoza2018d}. This unstable setting and the fact that researchers were not paying sufficient attention to reproducibility led to the same problems being solved over and over again \cite{Pacha2018d}.
\item \emph{Lack of standards representations}: There exist no standard representation formats for OMR outputs, especially not for structured encoding, and virtually every system comes with its own internal representation and output format, even for intermediate steps. This causes incompatibilities between different systems and makes it very hard to replace subcomponents. Work on underlying formalisms for describing music notation can also potentially have a wide impact, especially if done in collaboration with the relevant communities (W3C Community Group on Music Notation, Music Encoding Initiative).
\item \emph{Evaluation}: Due to the lack of standards for outputting OMR results, evaluating them is currently in an equally unsatisfactory state. An ideal evaluation method would be rigorously described and verified, have a public implementation, give meaningful results, and not rely on a particular use-case, thus only intrinsically evaluating the system \cite{Hajicjr.2018b}.
\end{itemize}

On the technical side, there are also many interesting avenues, where future research is needed, including:
\begin{itemize}
\item \emph{Music Object Detection}: recent work has shown that the music object detection stage can be addressed in one step with deep neural networks. However, the accuracy is still far from optimal, which is especially detrimental to the following stages of the pipeline that are based on these results. In order to improve the detection performance, it might be interesting to develop models that are specific to the type of inputs that OMR works on: large images with a high quantity of densely packed objects of various sizes from a vast vocabulary.
\item \emph{Semantic reconstruction}: merely detecting the music objects in the document does not represent a complete music notation recognition system, and so the music object detection stage must be complemented with the semantic reconstruction. Traditionally, this stage is addressed by hand-crafted heuristics that either hardly generalize or do not cover the full spectrum of music notation. Machine learning-based semantic reconstruction represents an unexplored line of research that deserves further consideration.
\item \emph{Structured encoding research}: despite being the main motivation for OMR in many cases, there is a lack of scientific research and open systems that actually pursue the objective of reconstructing the full structure encoding of the input.
\item \emph{Full end-to-end systems}: end-to-end systems are accountable for major advances in machine learning tasks such as text recognition, speech recognition, or machine translation. The state of the art of these fields is based on recurrent neural networks. For design reasons, these networks currently deal only with one-dimensional output sequences. This fits the aforementioned tasks quite naturally since their outputs are mainly composed of word sequences. However, its application for music notation---except for simple monophonic scores---is not so straightforward, and it is unknown how to formulate an end-to-end learning process for the recognition of fully-fledged music notation in documents.
\item \emph{Statistical modeling}: most machine learning algorithms are based on statistical models that are able to provide a probability distribution over the set of possible recognition hypotheses. When it comes to recognizing, we are typically interested in the best hypothesis---the one that is proposed as an answer---forgetting the probability given to such hypothesis by the model. However, it could be interesting to be able to exploit this uncertainty. For example, in the standard decomposition of stages in OMR systems, the semantic reconstruction stage could benefit from having a set of hypotheses about the objects detected in the previous stage, instead of single proposals. Then, the semantic reconstruction algorithm could establish relationships that are more logical {\it a priori}, although the objects involved have a lower probability according to the object detector. These types of approaches have not been deeply explored in the OMR field. Statistical modeling could also be useful so that the system provides a measure of its confidence about the output. Then, the end user might have a certain notion about the accuracy that has been obtained for the given input. This would be especially useful in an interactive system (see below).
\item \emph{Generalizing systems}: A pressing issue is generalizing from training datasets to various real-world collections because the costs for data acquisition are still significant and currently represent a bottleneck for applying state-of-the-art machine learning models in stakeholders' workflows. However, music notation follows the same underlying rules, regardless of graphical differences such as whether it is typeset or handwritten. Can one leverage a typeset sheet music dataset to train for handwritten notation? Given that typeset notation can be synthetically generated, this would open several opportunities to train handwritten systems without the effort of getting labeled data manually. Although it seems more difficult to transfer knowledge across different kinds of music notation, a system that recognizes some specific music notation could be somehow useful for the recognition of shared elements in other styles as well, e.g., across the various mensural notation systems.
\item \emph{Interactive systems}: Interactive systems are based on the idea of including users in the recognition process, given that they are necessary if there is no tolerance for errors---something that at the moment can only be ensured by human verification. This paradigm reformulates the objective of the system, which is no longer improving accuracy but reducing the effort---usually measured as time---that the users invest in aiding the machine to achieve that perfect result. This aid can be provided in many different ways: error corrections that then feed back into the system, or manually activating and deactivating constraints on the content to be recognized. However, since user effort is the most valuable resource, there is still a need to reformulate the problem based on this concept, which also includes aspects related to human-computer interfaces. The conventional interfaces of computers are designed for entering text (keyboard) or performing very specific actions (mouse); therefore, it would be interesting to study the use of more ergonomic interfaces to work with musical notation, such as an electronic pen or a MIDI piano, in the context of interactive OMR systems.
\end{itemize}

We hope that these lists demonstrate that OMR still provides many interesting challenges that await future research.

\bibliographystyle{plain}
\bibliography{UnderstandingOmr}

\newpage

\section*{Appendix A: OMR Bibliography}
\label{sec:AppendixA}

Along with this paper, we are also publishing the most comprehensive and complete bibliography on OMR that we were able to compile at \url{https://omr-research.github.io/}. It is a curated list of verified publications in an open-source Github repository (\url{https://github.com/OMR-Research/omr-research.github.io}) that is open for submissions both via pull requests and via templated issues. The website is automatically generated from the underlying BibTeX files using the BibTex2HTML library, available at \url{https://www.lri.fr/~filliatr/bibtex2html/}.

The repository contains three distinct bibliographic files that are rendered into separate pages:

\begin{enumerate}
\item \emph{OMR Research Bibliography}: A collection of scientific and technical publications, whose bibliographical metadata were manually verified for correctness from a trustworthy source (see below). Most of these entries have either a Digital Object Identifier (DOI) or a link to the website, where the publication can be found.
\item \emph{OMR Related Bibliography}: A collection of scientific and technical publications, whose bibliographical metadata were manually verified for correctness from a trustworthy source but are not primarily directed towards OMR, such as musicological research or general computer vision papers.
\item \emph{Unverified OMR Bibliography}: A collection of scientific and technical publications, that are related to Optical Music Recognition, but they could not be verified from a trustworthy source and might contain incorrect information. Many publications from this collection were authored before 1990 and are often not indexed by the search engines, or the respective proceedings could no longer be accessed and verified by us.
\end{enumerate}

\subsection*{Acquisition and Verification Process}
The bibliography was acquired and merged from multiple sources, such as the public and private collections from multiple researchers that have historically grown, including a recent one by Andrew Hankinson, who provided us with an extensive BibTeX library. Additionally, we have a Google Scholar Alert on the survey published by Rebelo et al. \cite{Rebelo2012}, as it currently represents the latest state-of-the-art review and is cited by almost every new publication.
To verify the information of each entry in the bibliography, we proceeded with the following steps:
\begin{enumerate}
\item Search on Google Scholar for the title of the work, if necessary with the authors last name and the year of publication.
\item Find a trustworthy source such as the original publisher, the authors' website, the website of the venue (that lists the article in the program) or indexing services including IEEE Xplore Digital Library, ACM Digital Library, Springer Link, Elsevier ScienceDirect, arXiv.org, dblp.org or ResearchGate. Information from the last three services are used with caution and if possible backed up with information from other sources.
\item Manually verify the correctness of the metadata by inspecting and correct it by obtaining the necessary information from another source, e.g., the conference website or the information state in the document. Suspicious information could be if the author's name is missing letters because of special characters or if the year of publication is before that of cited references.
\end{enumerate}

Once we verified the entry, we add it to the respective bibliography with JabRef (\url{http://www.jabref.org/}) and link the original PDF file or at least the DOI. Articles that were only found as PDF without the associated venue of publication were classified as technical reports. Bachelor theses and online sources such as websites of commercial applications were classified as 'Misc' because of the lack of an appropriate category in BibTex.

\end{document}